\documentclass[onefignum,onetabnum]{siamart220329}

\usepackage{lipsum}
\usepackage{amsfonts}
\usepackage{graphicx}
\usepackage{epstopdf}
\usepackage{algorithmic}
\usepackage{comment}

\ifpdf
  \DeclareGraphicsExtensions{.eps,.pdf,.png,.jpg}
\else
  \DeclareGraphicsExtensions{.eps}
\fi

\usepackage{dsfont}

\definecolor{red2}{RGB}{204,0,0}
\definecolor{blue2}{RGB}{0,103,165}
\hypersetup{colorlinks,linkcolor=[RGB]{0,103,165},citecolor=[RGB]{180,0,0}}
\newcommand{\RV}[1]{{\color{black} #1}}

\newsiamremark{remark}{Remark}
\newsiamremark{hypothesis}{Hypothesis}
\crefname{hypothesis}{Hypothesis}{Hypotheses}
\newsiamthm{claim}{Claim}

\def\be{\begin{equation}}
\def\ee{\end{equation}}

\def\x{\mathbf{x}}

\def\xh{\widehat{\x}}
\def\xm{\widetilde{\x}}
\def\y{\mathbf{y}}

\def\u{\mathbf{u}}
\def\f{\mathbf{f}}
\def\g{\mathbf{g}}
\def\z{\mathbf{z}}

\def\Z{\mathbf{Z}}
\def\Y{\mathbf{Y}}

\def\R{\mathbb{R}}
\def\Rs{\mathbb{R}}

\def\G{\mathbf{G}}

\def\Gh{\mathbf{\widehat{G}}}
\def\N{\mathbf{N}}

\def\T{\mathbf{T}}
\def\Pi{\mathbf{\Phi}}
\def\PPh{\mathbf{\Phi}}

\def\a{\boldsymbol{\alpha}}

\def\D{\mathbf{D}}

\def\Gam{\mathbf{\Gamma}}

\newcommand{\argmin}{\operatornamewithlimits{argmin}}
\newcommand{\argmax}{\operatornamewithlimits{argmax}}

\headers{Learning Nonautonomous SDE}{Yuan Chen and Dongbin Xiu}

\title{Modeling Unknown Stochastic Dynamical System Subject to
  External Excitation}

\author{Yuan Chen\and Dongbin Xiu\thanks{E-mail addresses: \texttt{\{chen.11050, xiu.16\}@osu.edu}. Department of Mathematics, The Ohio State University, Columbus, OH 43210, USA. Funding: This work was partially supported by AFOSR FA9550-22-1-0011.}}

\usepackage{amsopn}

\begin{document}

\maketitle

% REQUIRED
\begin{abstract}
We present a numerical method for learning an unknown nonautonomous stochastic dynamical system\RV{s}, i.e., stochastic system\RV{s} subject to time dependent
excitation or control signals. Our basic assumption is that the governing equations for the stochastic system are unavailable. However, short bursts of input/output (I/O) data consisting of certain known excitation signals and their corresponding system responses are available. 
When a sufficient amount of such I/O data are available, our method is capable of learning the unknown dynamics and producing an accurate predictive
model for the stochastic responses of the system subject to arbitrary excitation signals not in the training data.  Our method has two key components: (1) a local approximation of the training I/O data to transfer the learning into a parameterized form; and (2) a generative model to approximate the underlying
unknown stochastic flow map in distribution. After presenting the method in detail, we present a comprehensive set of numerical examples to demonstrate the performance of the proposed  method, especially for long-term system predictions.

\end{abstract}

% REQUIRED

% REQUIRED

\section{Introduction}
There has been a growing interest in recovering/discovering unknown dynamical systems from observational data.
Most of the existing studies focus on deterministic systems, with methods such as physics-informed neural networks (PINNs)
\cite{raissi2019physics,raissi2018multistep},  SINDy \cite{brunton2016discovering}, Fourier neural operator (FNO) \cite{li2020fourier}, computational graph completion \cite{owhadi2021computational}, sparsity promoting methods \cite{schaeffer2017sparse,schaeffer2018extracting,kang2019identifying}, flow map learning (FML) \cite{qin2019data, Churchill_2023}, to name a few.

Learning unknown stochastic systems is notably more challenging, as the stochastic noises in the systems usually can not be directly observed. The existing work
utilizes Gaussian process \cite{yildiz2018learning,pmlr-v1-archambeau07a,darcy2022one,opper2019variational}, polynomial approximations \cite{wang2022data, li2021data}, deep neural networks (DNNs) \cite{chen2023data,yang2022generative,chen2021solving,zhang2022multiauto, dietrich2023learning, zhu2024dyngma}, \RV{Koopman operators \cite{colbrook2024beyond,mezic2004comparison}}, etc. More recently, a stochastic extension of the deterministic flow map learning (FML) approach \cite{qin2019data, Churchill_2023} was proposed. It employs generative models such as GANs (generative adversarial networks) \cite{chen2023learning} or autoencoders \cite{xu2023learning} to model the underlying stochasticity. 
However,  most, if not all, of these methods are developed for autonomous systems, where time-invariance (in distribution) holds true and is critical to the method development.

The focus and contribution of this paper is on the learning and modeling of unknown non-autonomous stochastic systems. More specifically, we consider SDEs with unknown governing equations and subject to time dependent external excitation or control signals.
Our goal is to develop a method that can capture the stochastic dynamics of the unknown systems by using short-term data consisting of input/output (I/O) relations between the
excitation signals and their corresponding system responses. We remark that there exist some studies on modeling deterministic non-autonomous systems, using
methodology such as \RV{Dynamic Mode Decomposition (DMD) \cite{lu2023data,proctor2016dynamic,korda2018linear,kaiser2018sparse})}, SINDy (\cite{brunton2016sparse}, \RV{Koopman operator \cite{proctor2018generalizing,mauroy2019koopman,otto2021koopman}}, FML \cite{qin2021data}, etc. These methods are not applicable for stochastic non-autonomous systems.

The proposed method in this paper has two key components. First, our method utilizes the observational I/O data to construct an accurate representation of unknown stochastic dynamics of the system. This is accomplished by a generative model that learns the stochastic mapping of the system between two consecutive discrete time steps. The learning of this stochastic flow map is similar to the work of \cite{chen2023learning,xu2023learning}, which extended the deterministic FML to stochastic systems. While \cite{chen2023learning,xu2023learning} utilized GANs and autoencoder as the generative model, in this paper we employ conditional normalizing flow (cf. \cite{papamakarios2021normalizing}).  Normalizing flow has been widely adopted as a probabilistic model for generating data with desired distributions. Its applications include image and video generation \cite{laparra2011iterative}, statistical inference and sampling \cite{muller2019neural,song2017nice}, reinforcement learning \cite{haarnoja2018latent}, as well as scientific computing \cite{lu2022learning,li2022extracting,guo2022normalizing,deng2020modeling}. \RV{It has also been used to model temporal evolution of density function for stochastic system \cite{YuEtal_2022}. In this paper, we demonstrate that normalizing flow is applicable to modeling unknown stochastic systems and can serve as an effective alternative.}
The second key component of the proposed method is local parameterization of the excitation signal in the training I/O data. The local parameterization approach was first introduced in \cite{qin2021data} for deterministic nonautonomous sytem. We adopt the similar idea and extend it to stochastic system. The approach seeks to parameterize the excitation sigals in the training data via a local polynomial over one time step, \RV{and subsequently transform the non-autonomous stochastic system into a local Markovian system with homogeneous increments, provided that the unknown stochastic input has stationary and
independent increments.} This in turn transforms the learning problem into a parametric learning between the coefficients of the local polynomials and the system responses. This is a critical component, as it allows the learned system to conduct long-term system predictions under arbitrary excitation signals that are never seen in the training data.
Although the proposed method requires a large number of short bursts data, the overall demand for \RV{long trajectory} data may not be as large, because each burst of the training I/O data consist of merely two time steps of data entries. Once trained, the learned model is able to simulate the unknown stochastic systems for very long time and subject to arbitrary exitation/control signals. \RV{A prominent feature of the method, perhaps unique to most of the existing methods, is that it does not require the unknown system to be a classical SDE (with drift and diffusion) and is applicable to learning general non-Gaussian stochastic systems.}
\section{Setup}\label{sec:setup}

Let $\Omega$ be an event space and $T$ a finite time horizon.  \RV{We
consider a $d$-dimensional ($d\geq 1$) stochastic process $\x(\omega,
t): \Omega \times [0,T] \mapsto \mathbb{R}^d$ driven by an unknown
non-autonomous stochastic dynamical system in the following general
form
\begin{equation}
\label{eq:general}
    \frac{d \mathbf{x}_t}{dt}(\omega) = \mathbf{f}(\mathbf{x}_t,
    \u(t), \omega), 
  \end{equation}
 where $\omega\in\Omega$ represents the random input and $\u(t)$ are external
  excitation or control signals. Our basic assumption in this paper is
  that the governing equations of the stochastic system are not available. That is,
the right-hand-side (RHS) 
$\mathbf{f}$ is unknown. Consequently, the solution $\x_t$ can not be obtained by
solving \eqref{eq:general}.}

    \RV{We assume obervation data are available. More specifically,} we have input-output (I/O) time history data between the excitations $\u$, i.e., the inputs, and system response $\x$, i.e., the output,
    \be \label{io}
    \textrm{I/O training data:}\qquad \u(t) \rightarrow \x(t).
    \ee
    Our goal is to construct a numerical model for the unknown system \eqref{eq:general} such that it can produce accurate predictions of the
    system response $\x(t)$ for arbitrarily given excitations $\u(t)$
    that are not observed in the training data \eqref{io}.

%\be
%\label{equ:gSDE1}
%    d \x_t = \mathbf{f}(\mathbf{x}_t,t)dt + \boldsymbol{\sigma}(\mathbf{x}_t,t)d\mathbf{W}_t, \qquad \x_{t_0}(\omega)\sim \P_0,
%\ee
%where $\mathbf{W}_t$ is $m$-dimensional Brownian motion, and the
%unknown function $\mathbf{f}:\Rs^d\times \Rs\to\Rs^d$,
%$\boldsymbol{\sigma}:\Rs^d\times \Rs\to\Rs^{d\times m}$, $m\geq 1$,
%satisfy appropriate conditions (e.g., Lipschitz continuity). Without
%loss of generality, we consider the following form of SDE with
%control
\RV{
{\em Example 1:}
  As an example of the unknown stochastic system \eqref{eq:general}, let us consider a
  non-autonomous stochastic differential equation (SDE)}
\be
\label{equ:gSDE}
    d \x_t = \mathbf{a}(\mathbf{x}_t,\mu(t))dt + \mathbf{b}(\mathbf{x}_t,\nu(t))d\mathbf{W}_t,
    \ee
    where $\mathbf{W}_t$ is $m$-dimensional ($m\geq 1$) Brownian
    motion, $\mathbf{a}:\Rs^d\times \Rs\to\Rs^d$ drift function,
    $\mathbf{b}:\Rs^d\times \Rs\to\Rs^{d\times m}$ diffusion function,
    and 
    $\mu(t)$ and $\nu(t)$ time-dependent external inputs into the
    stochastic system.
    \RV{In our setting, both the drift $\mathbf{a}$ and the
    diffusion $\mathbf{b}$ are unknown, and the excitation signal is
    $\u(t) = (\mu(t), \nu(t))^T$.}
    %For notational convenience and without loss of generality, we will also write them as scalar functions.
    %Our method can be naturally extended to the general case in a component sense.
%
  %  Our basic assumption is that the SDE is unknown, in the sense that
   % the functions $\mathbf{a}$ and $\mathbf{b}$ are not known.
    Also, the
    driving Brownian motion $\mathbf{W}_t$ can not be
    observed. \RV{We emphasize that even though many stochastic system
      can be modeled in the form of \eqref{equ:gSDE}, our proposed
      method does not require such a structure. In fact, the proposed
      method works for unknown stochastic system of the general form
      \eqref{eq:general}, where the random inputs do not need to be
      Gaussian or Brownian motion.
  }

\RV{
  {\em Example 2:}
  Another example is reaction network, which can be modeled as
  stochastic jump process. For example, consider 
  a gene expression model in the form of biochemical reaction network \cite{voliotis2016stochastic,bowsher2013fidelity},
\begin{equation} \label{gene}
    \begin{split}
        \emptyset     & \xrightarrow{k(t)} M, \\
        M & \xrightarrow{k_\text{s}} M+P, \\
        M & \xrightarrow{k_{\text{dm}}} \emptyset, \\
        P & \xrightarrow{k_{\text{dp}}} \emptyset,
    \end{split}
  \end{equation}
where $M$ is mRNA population and $P$ is protein population, and paramters $k$'s are
reaction rates. Due to external excitation,
one of the reaction rates becomes time dependent $k(t)$, whereas the
other reaction rates remain constant. 
The dynamical behavior of $M$ and $P$ can be modeled by an
exact Monte Carlo method known as Gillespie algorithm, or stochastic
simulation algorithm (SSA), cf. \cite{gillespie1977exact}.
In our setting, we assume that the reaction network \eqref{gene} are not
known. Subsequently, it is not possible to conduct SSA to predict
the system behavior.
We assume we can collect time history data of $(M, P)$ subject to
certain known external signal $k(t)$.
That is, in the form of the I/O data \eqref{io}, the state
variables are $\x(t) = (M, P)^T$ and the excitation signal is $\u(t) =
k(t)$. Our goal is to model the long-term dynamics of $\x(t)$ as a
function of $k(t)$. Note that the stochastic process $\x(t)$ can not
be modeled by the SDE \eqref{equ:gSDE}. It is not
driven by Brownian motion and does not follow Gaussian
distribution. This will be one of the numerical examples in the paper.
}
  
    \subsection{Problem Statement}
The method presented in this paper is based on discrete time setting. 
    Let $t_0 < t_1<...$ be discrete time points. For simplicity, we
    assume the time steps are of uniform length $\Delta =
    t_{i+1}-t_i$, $\forall i\geq 0$.
    Suppose we observe $N_T \geq 1$ I/O sequences of solution responses subject to
    input excitations: for $i=1,\dots,N_T$,
\be \label{traj}
    \left(\u\left(t_0^{(i)}\right),\u\left(t_1^{(i)}\right),\RV{\ldots},\u\left(t_{L_i}^{(i)}\right)\right) \rightarrow
    \left(\x\left(t_0^{(i)}\right),\x\left(t_1^{(i)}\right),\RV{\ldots},\x\left(t_{L_i}^{(i)}\right)\right),
\ee
where $(L_i+1)$ is the length of the $i$-th observation sequence.
%The external processes $\mu(\cdot)$ and $\nu(\cdot)$ are known either analytically or discrete at these time instances, recorded as
%
%For simplicity, we assume the discrete times are equally spaced and
%each observation sequence is of the same length, i.e. $\Delta: =
%t_n^{(i)}-t_{n-1}^{(i)}$ and $L_i \equiv L \geq 1$ for any $n \geq 1$
%and $i=1,...,N_T$.
Note that each sequence of the I/O data can cover different
time spans. Also, one may have more information about the excitation
$\u(t)$ beyond its point values. For example, the analytical form of
$\u(t)$ may be known within the time interval $[t_0^{(i)},
t_{L_i}^{(i)}]$ for some sequences.
%An illustration of our dataset can be found in Figure \ref{fig:data}.

%\begin{figure}[htbp]
  %\centering
  %\includegraphics[width=.99\textwidth]{./SDEDataNonauto.pdf}
  %\caption{An illustration of the trajectory data set.}
  %\label{fig:data}
%\end{figure}
%\begin{remark}
%An important property of the proposed method is that it does not require data on
%the time instances $t_n^{(i)}$, for all $n$ and $i$. This reduces the effort for
%data acqusition. However, more importantly, this is a critical property that allows
%our method to conduct long-term system predictions beyond the time domain
%from the training data. How is this accomplished will be made clear in the next section on the details of the method.
%\end{remark}

The objective is to construct a numerical model to predict the system response of \eqref{equ:gSDE} subject to arbitrary excitations.
More specifically, given an initial condition $\x_0$ and excitation signal $\u(t)$ that is not in the training I/O data \eqref{traj}, we require
the model prediction $\xh$ to approximate the true system response $\x$, i.e.,
\be \label{distribution}
    \xh(t_n; \x_0,\u(t)) \stackrel{d}{\approx} \x(t_n; \x_0,\u(t)), \qquad n=1,\dots,
\ee
where $ \stackrel{d}{\approx}$ stands for approximation in
distribution. Note that since in general the stochastic driving term
$\mathbf{W}_t$ can not be directly observed, a weak approximation,
such as approximation in distribution, is
typically the most one can achieve from a mathematical point of
view. 

\subsection{Related Work and Contribution}

This method developed in this paper has its foundation in two recent
\RV{works}: flow map learning (FML) for modeling deterministic unknown dynamical
systems and its extension to modeling stochastic dynamical systems.

For an unknown deterministic autonomous system,
$\frac{d\x}{dt} = \f(\x)$, $\x\in\R^d$,
where $\f:\R^d\to \R^d$ is unknown\RV{, the} FML method seeks to approximate the unknown flow map
$ \x_n = \PPh_{t_n-t_s}(\x_s)$ by using observation data. More specifically, by using data on $\x$ over one time step $\Delta t$,
the FML method constructs a model
$$
\x_{n+1} = \widetilde{\PPh}_{\Delta t}(\x_n),  
$$
where $\widetilde{\PPh}_{\Delta t} \approx  {\PPh}_{\Delta t}$ is a
numerical approximation of the true flow map over one time step
$\Delta t$.
Once constructed, the FML model can be used as a time marching scheme
to predict the system response under a given initial condition.
This framework was
proposed in  \cite{qin2019data}, with extensions to partially observed
system \cite{FuChangXiu_JMLMC20}, parametric systems
\cite{QinChenJakemanXiu_IJUQ}, as well as non-autonomous deterministic
system \cite{qin2021data}.
%where a local parameterization of the
%external inputs is employed to transfer the problem into a parametric system.
%For a review for learning dynamical systems, see \cite{Churchill_2023}.
%
%For non-autonomous deterministic dynamical system subject to external excitations
%$
%\frac{d\x}{dt} = \f(\x,\u(t)),
%$
%a local parameterization method was proposed in to approximate the excitations over one time step. By doing so,
%the method transforms the unknown system into a parametric system, where the method from \cite{QinChenJakemanXiu_IJUQ} is applicable.

%\subsubsection{Learning Autonomous SDE}
For learning autonomous stochastic system,
$
\frac{d\x}{dt} = \f(\x,\omega(t)),
$
where $\omega(t)$ represents an unknown stochastic process driving the system. The work of
\cite{chen2023learning} developed stochastic flow map learning
(sFML). Assuming the system satisfies time-homogeneous property  (\cite{oksendal2003stochastic})
%	\begin{equation} \label{autonomous}
$
\mathbb{P}(\mathbf{x}_{s+\Delta
  t}|\mathbf{x}_{s})=\mathbb{P}(\mathbf{x}_{\Delta
  t}|\mathbf{x}_{0})$, $s\geq 0$,
% \end{equation}
the method uses the observation data on the state variable $\x$ to
construct a one-step generative model
$$
    \x _{n+1}=\G_{\Delta t}(\x_n; \z),
    $$
    where $\z$ is a random variable with known distribution (e.g.,
    standard Gaussian). The function $\G$, termed stochastic flow map,
    approximates the conditional distribution
$
\G_{\Delta t}(\mathbf{x}_{s}; \z) \approx
\mathbb{P}(\mathbf{x}_{s+\Delta t}|\mathbf{x}_{s}).
$
Subsequently, the sFML model becomes a weak approximation, in
distribution, to the true stochastic dynamics. Different generative
models can be employed under the sFML framework. For example, generative adversarial networks (GANs) are
used in \cite{chen2023learning}, and an autoencoder is employed in \cite{xu2023learning}.

The primary contribution of this paper is on the development of data
driven modeling for unknown stochastic systems subject to external
excitations.
To accomplish this, we extend the sFML framework
(\cite{chen2023learning}), which was developed for autonomous system,
to non-autonomous stochastic
system. To learn the system I/O responses, we employ the local
parameterization technique developed for non-autonomous deterministic
system (\cite{qin2021data}). The method parameterizes the input
excitations in the data and transforms the learning problem
into learning a parametric dynamical system. For stochastic
non-autonomous system considered in this paper, we incorporate the method into
a generative model in the sFML framework. \RV{In particular, we use
normalizing flow as the generative model, which has not been
considered in stochastic flow map learning.} We shall demonstrate that the
newly developed method is highly effective in modeling unknown
stochastic systems, when excitations are not present in the training data.

\section{Methodology}\label{sec:method}

In this section, we describe the proposed learning method in
detail. \RV{Since it is difficult to present the mathematical content in
the general form \eqref{eq:general}, without knowledge of the
equation and the stochastic inputs, our exposition in this section is
primarily base on the more concrete SDE setting \eqref{equ:gSDE}. It
shall be clear that the proposed method does not rely on the special structure
provided by \eqref{equ:gSDE}.}

\subsection{Parameterization of Inputs} \label{sec:localp}

%In this section, we extend the local parameterization procedure for deterministic problem (\cite{qin2021data}) to stochastic system. More importantly, we discuss how the procedure allows one to removal the time variable in the method development.  

\RV{Again, for the convenience of mathematical exposition, let} us consider
the unknown SDE \eqref{equ:gSDE} over time interval $[t_n, t_{n+1}]$, $n\geq 0$,
%\be 
%\x_{\Delta} = \x_0 + \int_0^\Delta \mathbf{a}(\x_s,\mu(s)) ds + \int_0^\Delta \mathbf{b}(\x_s,\nu(s)) d\mathbf{W}_s. 
%\ee
%Then we consider the evolution of the system on discrete time points 

%One can have, for $n=0,1,...,L-1$
\be \label{x_n}
\x(t_{n+1}) = \x({t_n}) + \int_{t_n}^{t_{n+1}} \mathbf{a}(\x(s),\mu(s)) ds + \int_{t_n}^{t_{n+1}} \mathbf{b}(\x(s),\nu(s)) d\mathbf{W}(s),
\ee
which can be \RV{written} equivalently as,
\be
\begin{split} \label{ustep}
  \x({t_{n+1}}) = \x({t_n}) &+ \int_{0}^{\Delta} \mathbf{a}(\x({t_n+\tau}),\mu(t_n+\tau)) d\tau \\
  &+ \int_{0}^{\Delta} \mathbf{b}(\x({t_n+\tau}),\nu(t_n+\tau)) d\mathbf{W}(t_n+\tau).
  \end{split}
\ee
By using the compact notation $\u(t)=(\mu(t), \nu(t))^T$, we now consider the excitation $\u(t)$ in the time interval $[t_n, t_{n+1}]$.
Given the information of the excitation in the training data \eqref{traj}, we construct a parameterized form
\be \label{uGama}
\u(t)|_{[t_n, t_{n+1})} \approx
\widetilde{\u}(\tau; \Gam_n) = \sum_{k=1}^{m} \a_n^k~p_k(\tau), \qquad
\tau\in [0,\Delta),
\ee
%To be specific, we define $\widetilde{\mu}_n$ and $\widetilde{\nu}_n$ such that for $\tau \in\left[0, \Delta\right]$
%\begin{align} \label{equ:localpara}
%    \widetilde{\mu}_n\left(\tau ; \boldsymbol{\Gam}_n\right):=\sum_{j=1}^{n_{\mu}} \widehat{\mu}_n^j \phi_j(\tau) \approx \mu\left(t_n+\tau\right), \\
%    \widetilde{\nu}_n\left(\tau ; \boldsymbol{\Gam}_n\right):=\sum_{j=1}^{n_{\nu}} \widehat{\nu}_n^j \psi_j(\tau) \approx \nu\left(t_n+\tau\right),
%\end{align}
where $\{p_k, k=1,\dots, m\}$ is a set of prescribed analytical basis functions and
\be
    \Gam_n = \{\a_n^1,\dots, \a_{n}^m\} \in \Rs^{n_\Gamma},
\ee
are the expansion coefficients. In principle, one can choose any
suitable basis functions. Since the time interval $[t_n, t_{n+1}]$
usually has a (very)
small step size
$\Delta$, it suffices to use low-order polynomials. In fact,
low-degree monomials
bases, $p_k(\tau) = \tau^{k-1}$, $k\geq 1$, would be sufficient for
most problems. When $k=0$, the parameterization takes form of
piecewise constant function; when $k=1$, piecewise linear
function. \RV{If higher degree polynomial approximation is needed, one
  should employ orthogonal polynomials as the basis for numerical stability.}

The local parameterization of $\u$ is carried out based on the information one has about the excitations. If the excitations are only known at
discrete time instances, as shown in \eqref{traj}, then it is natural to utilize piecewise linear \RV{function},
$$
\widetilde{\u}(\tau; \Gam_n) = \u(t_n) + \frac{\tau}{\Delta}(\u(t_{n+1}) - \u(t_n)).
$$
If more information about $\u(t)$ is available, one can construct a
higher degree polynomial.
We remark that in the representation, only the values of the
excitations $\u$ at $t_n$ and $t_{n+1}$ are needed. The values of the
time $t_n$ and $t_{n+1}$ are not required.

\RV{It will become clear in the following sections that the critical
  approximation in the proposed method is the approximation in distribution
  \eqref{distribution}, which is accomplished by a stochastic
  generative model via the observation data \eqref{traj}. Here, the
  numerical errors induced by randomness and finite sampling are
  dominant and overwhelm the small deterministic error of the local
  polynomial approximation. We
  have found no numerical advantage of using polynomials beyond second order.
}

%Using the local parameterization \eqref{equ:localpara}, we can define the global parameterized inputs along the time-axis
%\begin{align} \label{equ:globalpara}
%    \widetilde{\mu}(t ; \boldsymbol{\Gamma})=\sum_{n=0}^{N-1} \widetilde{\mu}_n\left(t-t_n ; \boldsymbol{\Gamma}_n\right) \mathds{1}_{\left[t_n, t_{n+1}\right]}(t), \\
%    \widetilde{\nu}(t ; \boldsymbol{\Gamma})=\sum_{n=0}^{N-1} \widetilde{\nu}_n\left(t-t_n ; \boldsymbol{\Gamma}_n\right) \mathds{1}_{\left[t_n, t_{n+1}\right]}(t),
%\end{align}
%where 
%\be
%    \boldsymbol{\Gamma}=\left\{\boldsymbol{\Gamma}_n\right\}_{n=0}^{L-1} \in \mathbb{R}^{L \times (n_\mu+n_\nu)}.
%\ee
%Then we define a modified SDE system of \eqref{equ:gSDE1}
%\be
%\label{equ:gSDEmodify}
%    d \xm_t = \mathbf{a}\left(\xm_t,\widetilde{\mu}(t ; \boldsymbol{\Gamma})\right)dt + \mathbf{b}\left(\xm_t,\widetilde{\nu}(t ; %\boldsymbol{\Gamma})\right)d\mathbf{W}_t.
%\ee
%This system is an approximation of system \eqref{equ:gSDE1}. The approximation capability depends on the accuracy of $\widetilde{\mu}(\cdot)$ and $\widetilde{\nu}(\cdot)$ to the analytical inputs $\mu(\cdot)$ and $\nu(\cdot)$. 

\subsection{Parametric Stochastic Flow Map}

By replacing $\u$ by the local polynomial $\widetilde{\u}$
\eqref{uGama}, we transform the system \eqref{ustep} into
\be \label{xx_n}
\begin{split}
  \xm({t_{n+1}}) = \xm({t_n}) &+ \int_{0}^{\Delta} \mathbf{a}(\xm({t_n+\tau}),\widetilde{\mu}\left(\tau ; {\Gam}_n\right)) d\tau \\
  & + \int_{0}^{\Delta} \mathbf{b}(\xm({t_n+\tau}),\widetilde{\nu}\left(\tau ; {\Gam}_n\right)) d\mathbf{W}(t_n+\tau),
  \end{split}
\ee
%Like the original system \eqref{x_n}, the functions $\mathbf{a}$ and
%$\mathbf{b}$, along with the driving stochastic process $\mathbf{W}$,
%are unknown.
where the
excitation signals $\u=(\mu, \nu)^T$ has been parameterized by
$\widetilde{\u}$ via a set of parameters $\Gam_n$. Compared to
\eqref{ustep}, the transformed system \eqref{xx_n} contains possible numerical
error introduced by the parameterization of the excitations over the
time domain $[0,\Delta)$. The error can be made arbitrarily small \RV{for smooth excitations} if
one uses higher degree polynomials when $\Delta$ is sufficiently small.

By using \RV{subscripts} to denote the time level and letting
$$
\xm(t_n) = \xm_n, \qquad d\mathbf{W}_n(\tau) = d\mathbf{W}(t_n+\tau),
$$
the parameterized system \eqref{xx_n} indicates that there exists a
mapping
\be \label{sFM}
    \xm_{n+1} = \G_\Delta(\xm_{n},d\mathbf{W}_n(\Delta); \boldsymbol{\Gamma}_{n}),
    \ee
    where $\G_\Delta$
    is what we shall call {\em
  parametric stochastic flow map}, which is parameterized by
$\Gam_n$. It is an unknown operator as the
functions $\mathbf{a}$ and $\mathbf{b}$ are unknown in the original
system \eqref{equ:gSDE}.

\begin{remark} \label{remark1}
It is important to recognize that for the Brownian motion
$\mathbf{W}(t)$, or in general for L\'{e}vy processes (c\`{a}dl\`{a}g
stochastic processes with stationary independent increments), 
the process $d\mathbf{W}_n(\tau)$ is stationary and independent of
$\mathbf{W}(t_n)$.  Therefore, only the time difference $\Delta =
t_{n+1} - t_n$ matters, and the values of $t_n$ and
$t_{n+1}$ do not.  Consequently, we have suppressed the time variable
$t_n$ and $t_{n+1}$ in \eqref{sFM}.
\end{remark}

\begin{remark} \label{remark2}
\RV{The derivation of \eqref{sFM}, via the suppression of the time
  variable $t_n$, represents a subtle and yet signficant step. The
  parametric stochastic flow map \eqref{sFM} is Markovian, in the
  sense that $\xm_{n+1}$ depends only on the system information at
  $t_n$, i.e., on $\xm_n$, the local parameters for the excitation signal
  $\boldsymbol{\Gamma}_n$, and the stationary local stochastic
  increment $d\mathbf{W}_n$.}

\end{remark}

\RV{
It is also clear that the derivation of
\eqref{sFM} does not rely on the special structure of
\eqref{equ:gSDE}, so long as the stochastic input has stationary and
independent increments. Therefore, we conclude that the parametric
stochastic flow map \eqref{sFM} is applicable to the general system
\eqref{eq:general} when the stochastic inputs are  L\'{e}vy processes.
}

\subsection{Stochastic Flow Map Learning}

In this section, we describe our main method of stochastic flow map
learning (sFML), which constructs a generative
model to approximate the stochastic flow map \eqref{sFM} by using the
trajectory data \eqref{traj}.

\subsubsection{Training Data}

To construct the training data set, we reorganize the original
training data set \eqref{traj} into pairs of consecutive time
instances, \RV{often referred to as snapshots}. Since for each of the $i$-th trajectory, $i=1,\dots, N_T$, we can extract $L_i$
such pairs, there are a total number of
$M=L_1+\cdots+L_{N_T}$ I/O data pairs from the data set \eqref{traj}:
\be 
  \left(\u\left(t_k^{(j)}\right),\u\left(t_{k+1}^{(j)}\right)\right) \rightarrow
    \left(\x\left(t_k^{(j)}\right),\x\left(t_{k+1}^{(j)}\right)\right),
    \qquad j=1,\dots, M.
    \ee

    Next, we perform the local parameterization procedure from Section
    \ref{sec:localp} to each pair of the input data
    $\left(\u\left(t_k^{(j)}\right),\u\left(t_{k+1}^{(j)}\right)\right)$
    and obtain its parameterization $\Gam_k^{(j)}$, $j=1,\dots,M$, in
    the form of \eqref{uGama}. We now have
    \be
    \Gam_k^{(j)} \rightarrow
    \left(\x\left(t_k^{(j)}\right),\x\left(t_{k+1}^{(j)}\right)\right),  \qquad j=1,\dots, M.
    \ee

    Since the values of the time variables do not matter, see Remark
    \ref{remark1}, we again suppress the time variables and write our
    training data set as
    \be \label{dataset}
    \mathcal{S}_M = 
\left\{\boldsymbol{\Gam}_0^{(j)}; \left(\x_0^{(j)},
    \x_{1}^{(j)}\right)\right\}_{j=1}^M, 
\ee
where $M$ is the total number of the parametric data pairs. In this
way, each $j$-th entry of the data set is a trajectory of
length two over one time step $\Delta$, starting with its ``initial condition'' at $\x_0^{(j)}$,
ending one step later at $\x_1^{(j)}$, and driven by a
known excitation parameterized by $\boldsymbol{\Gam}_0^{(j)}$ and an
unknown stationary stochastic process $d\mathbf{W}_n(\Delta)$.
%In other words, the training data set is re-arranged into $M$ numbers of short trajectories of only two entries. Each of the $i$-th trajectory, associated with a parameter set $\boldsymbol{\Gam}^{(i)}$, $i=1,\dots, M$, starts with an ``initial condition'' $\x_0^{(i)}$ and ends a single time step $\Delta$ later at $\x_1^{(i)}$. The dataset \eqref{equ:data2o} serves as our training data.

\subsubsection{Generative Model}

In stochastic flow map learning (sFML), we seek to approximate the parametric
stochastic flow map \eqref{sFM} via a recursive
generative model in the form of
\be \label{sFM_model}
\xh_{n+1} = \Gh_\Delta (\xh_n, \z_n; \Gam_n),
\ee
where $\z\in\Rs^{n_z}$ is a random variable of known
distribution. Again, since the stationary stochastic process
$d\mathbf{W}_n(\Delta)$ in \eqref{sFM} is not observed, the
constructed sFML model \eqref{sFM_model} is expected to be a weak
approximation of 
\eqref{sFM}, and in this particular case, approximation in distribution.

In order to construct the sFML model \eqref{sFM_model}, we execute the
model for one time step over $\Delta$,
\be \label{sFM1}
\xh_{1} = \Gh_\Delta (\xh_0, \z_0; \Gam_0),
\ee
and utilize the training data set \eqref{dataset} to learn the unknown
operator $\Gh$. Note that the
random variable $\z_0$ is not in the training date set. In practice, one chooses $\z_0$ with a known
distribution, typically a standard Gaussian, and a specified
dimension $n_z\geq 1$. The presence of the random variable $\z_0$
enables \eqref{sFM1} to be a stochastic generative model that can produce random
realizations. Several methods exist to construct stochastic
generative models, e.g., GANs, diffusion model, normalizing flow,
autoencoder-decoder, etc. In this paper, we adopt normalizing flow for
\eqref{sFM1}.
\RV{We remark that the use of normalizing flow is not critical to the
  applicability of the proposed method. Our numerical experiments have
  indicated that other generative models such as GANs and diffusion
  model work generally well.  We employ normalizing flow primarily because it has not been
  used in the FML framework before. The property, applicability and
  performance comparison of different generative models is a rather
  complex issue and
  out of the scope of this paper.}

\subsubsection{Normalizing Flow Model}

Normalizing flows are generative models that produce tractable
distributions to enable efficient and accurate sampling and density
evaluation. A normalizing flow is a transformation of a simple
probability distribution, e.g., a standard normal, into a more complex
distribution by a sequence of diffeomorphism.
Let $\Z\in\Rs^D$ be a random variable with a known and
tractable distribution $p_\Z$. Let $\g$ be a diffeomorphism, whose
inverse is $\f=\g^{-1}$, and $\Y=\g(\Z)$. Then using the change of
variable formula, one obtain the probability of $\Y$:
$$
p_\Y(\y)= p_\Z(\f(\y)) |\det \D\f(\y)|= p_\Z(\f(\y)) |\det \D\g(\RV{\mathbf{f}}(\y))|^{-1},
$$
where $\D\f(\y) = \partial\f/\partial \y$ is the Jacobian of $\f$ and
$\D\g(\z) = \partial \g/\partial \z$ is the Jacobian of $\g$. When the
target complex distribution $p_\Y$ is given, usually as a set of
samples of $\Y$, one chooses $\g$ from a parameterized family
$\g_\theta$, where the parameter $\theta$ is optimized to match the
target distribution. Also,
to circumvent the difficulty of constructing a complicated nonlinear
function $\g$, one utilizes a composition of (much) simpler
diffeomorphisms: $\g = \g_m\circ \g_{m-1}\circ\cdots\circ\g_1$. It can
be shown that $\g$ remains a diffeomorphism with its inverse $\f =
\f_1\circ\cdots\circ\f_{m-1}\circ\f_m$. There exist a large amount of
literature on normalizing flows. We refer interested reader to review
articles \cite{Kobyzev2021, Papamakarios21}.

%Let $\T_\theta: \mathbb{R}^d \mapsto \mathbb{R}^d$ be a class of invertible maps parameterized by $\theta \in\mathbb{R}^{n_T}$, $n_T>0$ is dimension of the parameter space. Normalizing flow seeks to optimize $\theta$ such that $\T_\theta(\z)$ approximates the distribution of training data, where $\z$ usually follows a prescribed distribution, e.g. standard Gaussian distribution. 

In our setting, we seek to construct the one-step generative model
\eqref{sFM1} by using the training data \eqref{dataset}. Let
$\z_0\in\Rs^d$ be a random variable with a known distribution. In our
approach, we choose $\z_0$ to be $d$-dimensional standard normal.
Let $\T_\theta$ be a diffeomorphism with a set of
parameters $\theta\in\Rs^{n_\theta}$. Our objective is to find $\theta$ such that
$\T_\theta(\z_0)$ follows the distribution of $\{\x_1^{(j)}\}_{j=1}^M$
in \eqref{dataset}.

Since the distribution of $\x_1$ clearly depends on $\x_0$ and
$\Gam_0$, we constraint the choice of $\theta$ to be a function of
$\x_0$ and $\Gam_0$. We define
\be \label{theta}
\theta = \N(\x_0, \Gam_0; \Theta),
\ee
where $\N$ is a DNN mapping with trainable hyperparameters
$\Theta$. No special DNN structure required, and we adopt the
straightforward fully connected feedforward DNN for $\N$. This
effectively defines
\be \label{x1}
\x_1 = \T_{\N(\x_0, \Gam_0; \Theta)}(\z_0),
\ee
where the diffeomorphism $\T$ is effectively parameterized by the trainable
hyperparameters $\Theta$ of the DNN. Let  $\mathbf{S}=\T^{-1}$ be the
inverse of $\T$. We have $\z_0 =
\mathbf{S}_{\N_\Delta(\x_0,\Gam_0 ; \mathbf{\Theta})} (\x_1)$. 

%For our training data \eqref{dataset}, it is noted that the distribution of $\x_1^{(j)}$ depends on $\x_0^{(j)}$ and $\Gam_0^{(j)}$. This motivates us to express $\theta$ by an operator $\N_\Delta:\mathbb{R}^{n_\Gamma+d} \mapsto \mathbb{R}^{n_T}$, which depends on its inputs $(\x_0^{(j)},\Gam_0^{(j)})$. We construct the operator $\N_\Delta$ using a simple feed-forward DNN with trainable parameter $\mathbf{\Theta}$. 

%Now we seek an invertible mapping $\T_{\N_\Delta}$ such that for any $(\x_0,\Gam_0)$
%\be
%    \x_1| \x_0,\Gam_0 \stackrel{d}{=} \T_{\N_\Delta(\x_0,\Gam_0 ; \mathbf{\Theta})}(\z_0).
%\ee
%Obviously, the conditional distribution of $\x_1 | \x_0,\Gam_0$
%depends on training parameter $\mathbf{\Theta}$.
The invertibility of $\T$ allows us to compute:
%we can obtain the conditional density of $\x_1 | \x_0,\Gam_0$ using the formula change of variables:
\be \label{equ:lh}
    p(\x_1 |
    \x_0,\Gam_0;\mathbf{\Theta})=p_{\z_0}\left(\mathbf{S}_{\N(\x_0,\Gam_0
        ; \Theta)}(\x_1)\right)\left|\det \D{\T_{\N(\x_0,\Gam_0 ;
          {\Theta})}}(\mathbf{S}_{\N(\x_0,\Gam_0
        ; \Theta)} (\x_1))\right|^{-1}. 
\ee
%In the above formula, we use $p_\x(\cdot)$ to represent the
%(conditional) probability density function of random variable $\x$,
%this subscription may be omitted without causing confusion. $J_\T$
%denotes the Jacobian matrix of operator $\T$ and $\operatorname{det}$
%is its determinant. Note that the density \eqref{equ:lh} is
%computable, with a concrete implementation of the invertible mapping
%$\T$.
%We train the conditional normalizing flow model by maximizing the
%expected log-likelihood function
The hyperparameters $\Theta$ are determined by maximizing the
expected log-likelihood, which is accomplished by minimizing its
negative as the loss
function,
$$
   \mathcal{L}({\Theta}):=-\mathbb{E}_{(\Gam_0,\x_0,\x_1)\sim p_{\mathtt{data}}} \log(p(\x_1|\x_0,\Gam_0;{\Theta})),
$$
where $p_{\mathtt{data}}$ is the distribution from the training data
set \eqref{dataset} and computed as
\be \label{loss}
    \mathcal{L}({\Theta}) = -\sum_{j=1}^M \log\left(p(\x_1^{(j)} | \x_0^{(j)},\Gam_0^{(j)};{\Theta})\right).
    \ee

Several designs for the invertible map $\T$ have been developed and
studied extensively in the literature. These include, for example, masked
autoregressive flow (MAF) \cite{NIPS2017_6c1da886}, real-valued
non-volume preserving (RealNVP) \cite{dinh2017density}, neural
ordinary differential equations (Neural ODE)
\cite{NEURIPS2018_69386f6b}, etc.
In this paper, we adopt the MAF approach, where the dimension of the
parameter $\theta$ in \eqref{theta} is set to be $n_\theta = 2d$,
where $d$ is the dimension of the dynamical system. For the technical
detail of MAF, see \cite{NIPS2017_6c1da886}.

\subsection{DNN Model Structure and System Prediction}

\begin{figure}[htbp]
  \centering
  \includegraphics[width=.8\textwidth]{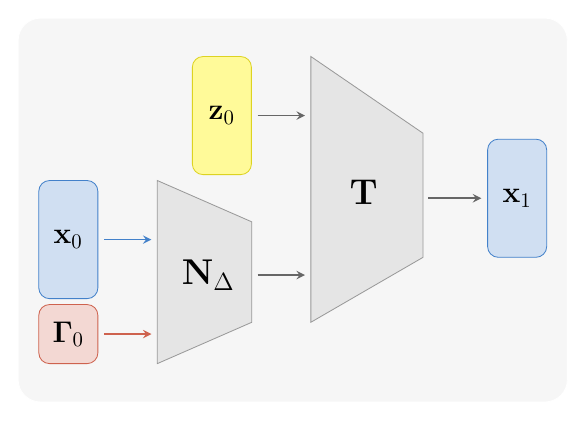}
  \caption{The DNN model structure for the proposed normalizing flow
    sFML method \eqref{x1}.}
  \label{fig:NFnet}
\end{figure}

An illustration of the proposed sFML model structure can be found in Figure
\ref{fig:NFnet}. This is in direct correspondence of \eqref{x1}.
Minimization of the loss function \eqref{loss}, using the data set
\eqref{dataset}, results in the training of the DNN hyperparameters
$\Theta$. Once the training is completed and $\Theta$ fixed,
\eqref{x1} effectively defines the one-step sFML model \eqref{sFM1}:
$$
\x_1 = \T_{\N(\x_0, \Gam_0)}(\z_0) = \Gh_\Delta (\x_0, \z_0; \Gam_0),
$$
where we have suppressed the fixed parameter $\Theta$.

Iterative execution of the one-step sFML model allows one to conduct
system predictions under excitations that are not in the training
data. For a given (new) excitation signal $\u(t) = (\mu(t),
\nu(t))^T$, we first conduct its parameterization in the form of
\eqref{uGama}, to obtain its local parameter $\Gam_n$ for $[t_n,
t_{n+1})$, for any $n\geq 0$. The sFML system then produces the system
prediction, for
a given initial condition $\x_0$,
\be \label{predict}
\x_{n+1} = \Gh_\Delta (\x_n, \z_n; \Gam_n), \qquad n\geq 0,
\ee
where $\z_n$ are i.i.d. $d$-dimensional standard normal random
variables.

\section{Numerical Examples}
\label{sec:examples}
In this section, we present several numerical tests to demonstrate the
performance of our proposed method. After presenting results for learning
an Ornstein-Uhlenbeck (OU) process and a nonlinear SDE, we focus on
nonlinear SDE systems for long-term predictions. 
These include stochastic a predator-prey model and a stochastic
oscillator with double well potential. In both cases, we study {\em
  very} long-term predictions of the learned sFML models. In
particular, for the stochastic oscillator, we utilize a periodic
excitation signal that is known to generate the well-known
``stochastic resonance'' phenomenon. \RV{At the end of this section,
  we present learning of a stochastic reaction network and a
  stochastic partial diferential equation (SPDE). The stochastic
  reaction network is a non-Gaussian stochastic jump process driven by
  its inherent stochasticity that is not Brownian motion. It
  demonstrates the applicability of the proposed method to general
  stochastic system \eqref{eq:general} beyond the classical SDE
  \eqref{equ:gSDE}. The SPDE example demonstrates the applicability of
  the method to a modestly high dimension $d=30$.}
% Our sFML model produces accurate prediction for as long as
% $T=40,000$.  

In all the examples, the true stochastic systems are known. However,
the true stochastic systems are used only to generate the training
data set \eqref{dataset}. When the true systems follow the SDE form
\eqref{equ:gSDE}, we solve them
by Euler-Maruyama method with a time step $\Delta=0.01$. The ``initial
conditions'' $\x_0$ in \eqref{dataset} are sampled uniformly in a
domain $I_{\x}$, specified in each example, and the excitations
are local polynomials whose coefficients $\Gam_0$ are sampled in a domain specified for each example.

%We use the Masked Autoregressive Flow (MAF) model as a technical realization of the normalizing flow.
%Upon use of the model, there are $3$ coupling flow mappings are constructed with independent training parameters. The training data size, batch size, and number of training epochs will be specified in each example.
In our sFML model, Figure
\ref{fig:NFnet},  the DNN $\N$ has 3 layers, each of which with 20 nodes, and utilizes $\tanh$ activation function. We employ cyclic learning rate with a base rate $3\times10^{-4}$ and a maximum rate $5\times10^{-4}$, $\gamma=0.99999$, and step size $10,000$.  The cycle is set for every $40,000$ training epochs and with a decay scale $0.5$.
A small weight decay of $0.01$ on the gradient updates is also used to help \RV{stabilize} the training. In our examples, the DNN training is usually conducted for $200,000\sim 300,000$ epochs.
%To reduce the overfitting phenomenon, we also add a small weight
%decay with a scale $0.01$ on the update gradients.

\RV{
For validation tests, we conduct system predictions by the
learned FML for time domains well beyond that of the training data
set. We employ mostly periodic excitation signals to ensure that the
long-term system behavior remains ``interesting'', that is, it does
not either decay to zero or blow up. We remark that periodicity of the
exciation is not required in the method, as it was never
assumed in the derivation of the method.}

\subsection{Linear SDE with Control}

We first consider Ornstein–Uhlenbeck (OU) process with control/excitation. Two cases are considered: when the control is in the drift and when the control is in both
the drift and the diffusion. Note that since the true equations are not known, one has no information on
``where'' the excitations operate onto the system. The sFML approach also does not seek to recover the drift or diffusion terms.

\subsubsection{OU with Drift Control}
\label{ex:OU1}
We first consider an Ornstein–Uhlenbeck (OU) process,
\begin{equation}
    dx_t = \left[-\mu x_t + \alpha(t) \right] dt+\sigma dW_t,
\end{equation}
where $\mu$ and $\sigma$ are set as $\mu=1.0$ and $\sigma=0.2$, and the control signal $\alpha(t)$ is applied to the drift.
The training data set \eqref{dataset} is generated by
sampling $x_0$ in $(-2,2)$ and using Taylor polynomial of degree $2$ for the control $\alpha(t)$. This introduces 3 parameters for $\Gam_0$, which are
sampled from $(-9,9)^3$. A total of $M=120,000$ trajectory pairs are used in the training data set \eqref{dataset}, where the time step $\Delta = 0.01$.
%An extra time-dependent input $\alpha(t)$ is locally parameterized with polynomials of degree $2$ added to the linear drift function. As a result, we have a local parameter set $\Gam_n \in \mathbb{R}^3$. The DNN used in the normalizing flow model is built with $3$ layers and $20$ nodes per layer. The model is trained using $120,000$ pair trajectory data simulated from initial conditions sampled with $I_\x = [-2,2]$ and $I_{\Gam}=[-9,9]^3$. The training epochs are taken to be $200,000$.

Once the sFML model \eqref{sFM_model} is trained, we conduct system prediction for up to $T=10.0$, which requires 1,000 time steps.
%The predicted results under an excitation $\alpha(t)=0.5\sin(6t)$ are shown in Figure \ref{fig:OU_drift_control1_sample}, where some sample solution realizations by the
%sFML model are shown on the right and some realizations from the true system shown on the left.
%The two sets of solutions appear visually in agreement.
\begin{figure}[htbp]
  \centering
  \label{fig:OU_drift_control1_sample}
  \includegraphics[width=.48\textwidth]{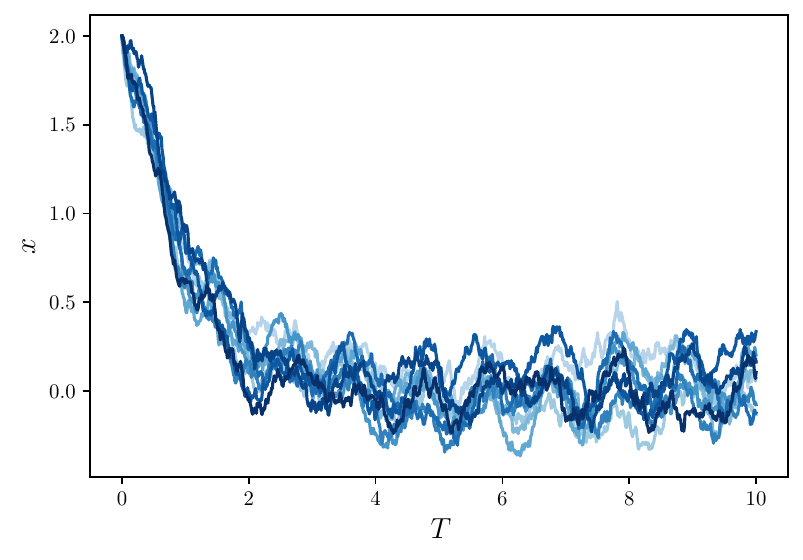}
  \includegraphics[width=.48\textwidth]{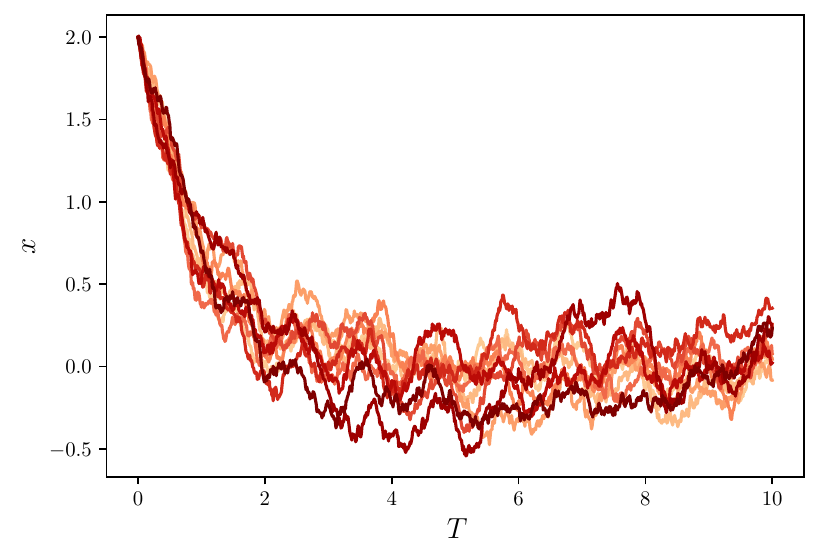}
  \caption{Sample trajectories of Example \ref{ex:OU1} with initial
    condition $x_0=2.0$ and $\alpha(t)=\frac{1}{2}\sin(6t)$. Left:
    ground truth; Right: Simulation using the trained sFML model. }
\end{figure}
\begin{figure}[htbp]
  \centering
  \label{fig:OU_drift_control1_ms}
  \includegraphics[width=.8\textwidth]{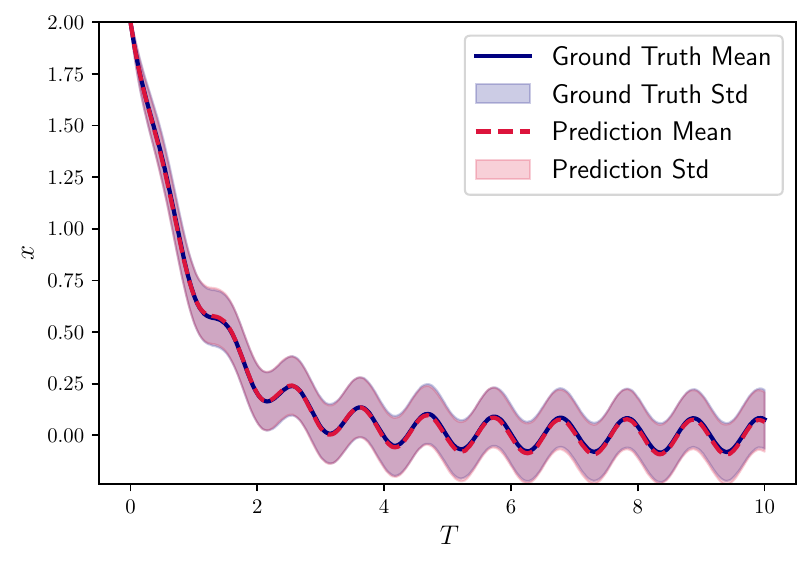}
  \caption{Mean and standard deviation (STD) of Example \ref{ex:OU1} with initial condition $x_0=2.0$ and $\alpha(t)=\frac{1}{2}\sin(6t)$.}
\end{figure}
\begin{figure}[htbp]
  \centering
  \label{fig:OU_drift_control1_pdf}
  \includegraphics[width=.32\textwidth]{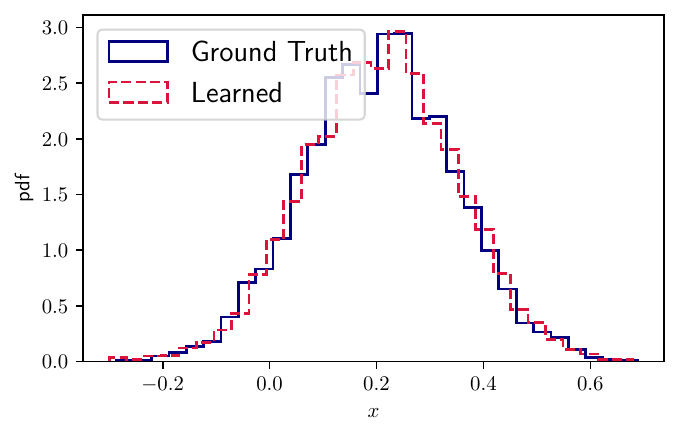}
  \includegraphics[width=.32\textwidth]{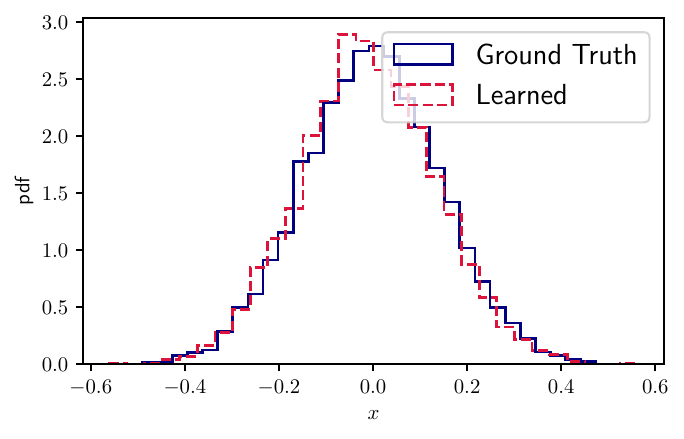}
  \includegraphics[width=.32\textwidth]{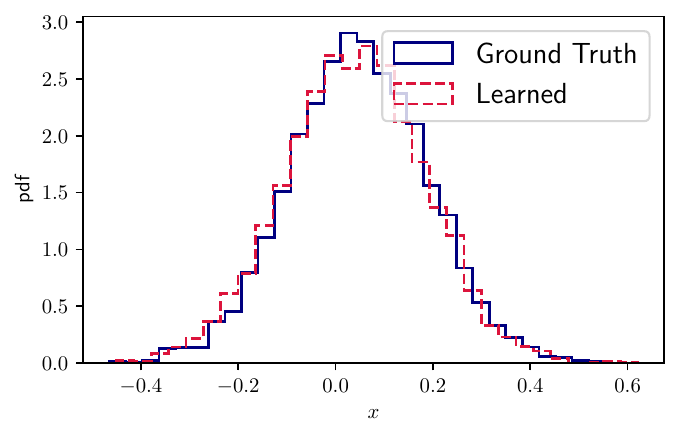}
  \caption{Comparasion of distribution of Example \ref{ex:OU1} at $T=2,4,8$ with initial condition $x_0=2.0$ and $\alpha(t)=0.5\sin(6t)$.}
\end{figure}
In Figure \ref{fig:OU_drift_control1_sample}, we compare some sample
trajectory \RV{paths} produced by the ground truth (left) and the learned
sFML model (right), with an initial condition $x_0=2.0$ and a ``new''
control signal $\alpha(t)=\frac{1}{2}\sin(6t)$. We observe the two
sets appear visually similar to each other. To further validate the
sFML model prediction, we compute the mean and standard deviation of
the solution averaged over $10,000$ trajectories. The sFML model
predictions are shown in Figure \ref{fig:OU_drift_control1_ms}, along
with the reference ground truth. \RV{In this figure, lines represent the mean of sample trajectories, while shadow bands depict one standard deviation above and below the mean curve. } In Figure
\ref{fig:OU_drift_control1_pdf}, we also show the comparison of the
solution probability distributions at time $T=2,4,8$. We observe good
agreement between the learned sFML model and the true model. This
verifies that the sFML model indeed provides an accurate approximation
in distribution.

\begin{figure}[htbp]
  \centering
  \label{fig:OU_drift_control2_sample}
  \includegraphics[width=.48\textwidth]{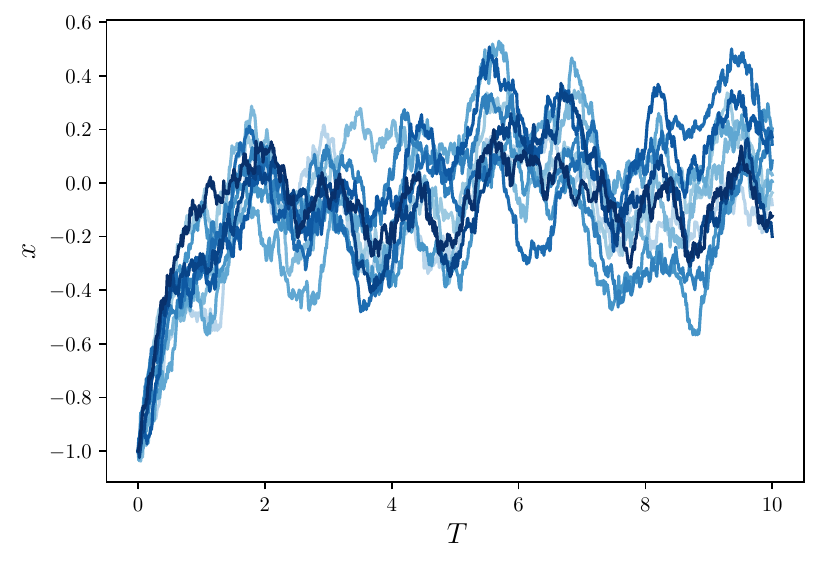}
  \includegraphics[width=.48\textwidth]{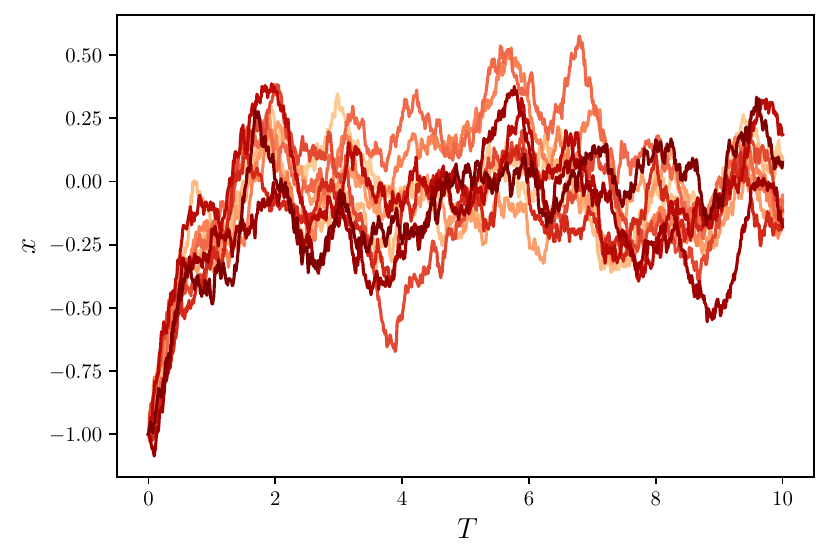}
  \caption{Sample trajectories of Example \ref{ex:OU1} with initial
    condition $x_0=-1.0$ and
    $\alpha(t)=\frac{1}{2}\sin(5t)+\frac{1}{5}\sin(1.5t)$. Left:
    ground truth; Right: Simulation using the trained sFML model.}
\end{figure}

\begin{figure}[htbp]
  \centering
  \label{fig:OU_drift_control2_ms}
  \includegraphics[width=.8\textwidth]{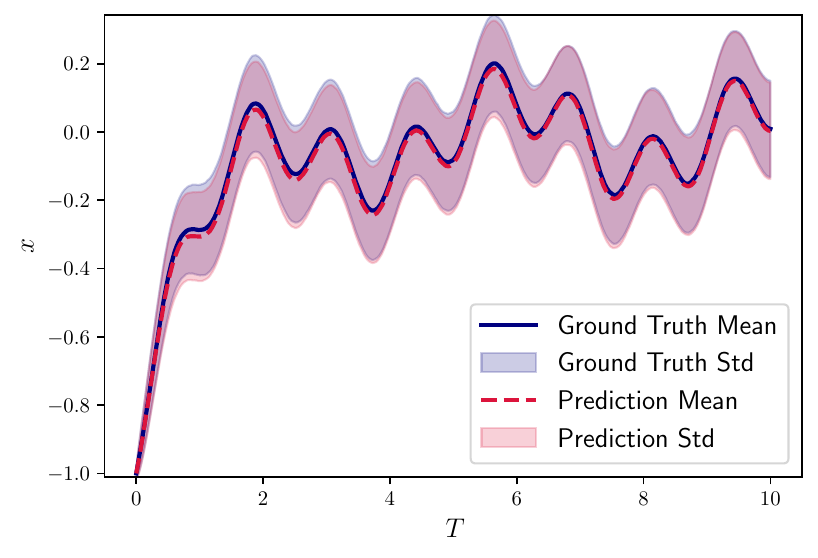}
  \caption{Mean and standard deviation (STD) of Example \ref{ex:OU1} with initial condition $x_0=-1.0$ and $\alpha(t)=\frac{1}{2}\sin(5t)+\frac{1}{5}\sin(1.5t)$.}
\end{figure}

We now present the results under a different setting: the initial
condition $x_0=-1.0$, and the excitation
$\alpha(t)=\frac{1}{2}\sin(5t)+\frac{1}{5}\sin(1.5t)$. The sample
solution trajectories are shown in Figure
\ref{fig:OU_drift_control2_sample} and the solution mean and standard
deviation averaged over $10,000$ trajectories are shown in Figure
\ref{fig:OU_drift_control2_ms}.  Again, we observe good agreement
between the sFML model prediction and the ground truth.  

\subsubsection{Fully control}
\label{ex:OU2}
% NonAutoEx19prod_a3b
We then consider the following OU process with control on both drift and diffusion terms:
\begin{equation}
    dx_t = \left[-\mu x_t + \alpha(t) \right] dt+\beta(t) dW_t,
\end{equation}
where $\mu=1.0$, and $\alpha(t)$ and $\beta(t)$ are the excitation/control signals. To generate training data, we conduct the local parameterization of  $\alpha(t)$ and $\beta(t)$ with 2nd degree Taylor polynomials, resulting in $\Gam_n \in \mathbb{R}^{n_\Gamma}$, $n_\Gamma=3+3=6$. Moreover, we generate $120,000$ training data pairs with initial conditions uniformly sampled from $I_\x=[-0.8,1.5]$ and $ I_\Gam = [-0.6,0.6]\times [-0.8,0.8]\times [-0.7,0.7]\times [0.01,0.35] \times [-0.5,0.5] \times [-1.55,0.55]$.
%The DNN used in our model is set with $3$ layers and $20$ nodes per layer. We train the model with $200,000$ epochs.
%
\begin{figure}[htbp]
  \centering
  \label{fig:OU_full_control1_sample}
  \includegraphics[width=.48\textwidth]{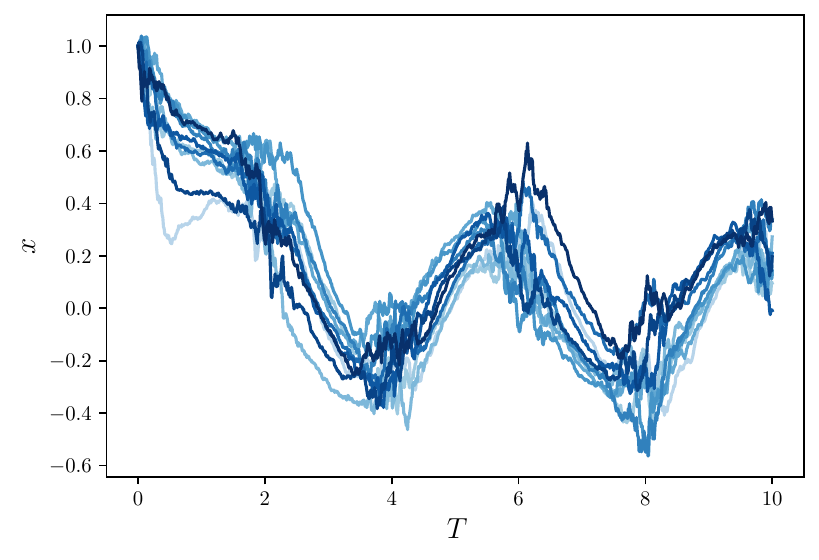}
  \includegraphics[width=.48\textwidth]{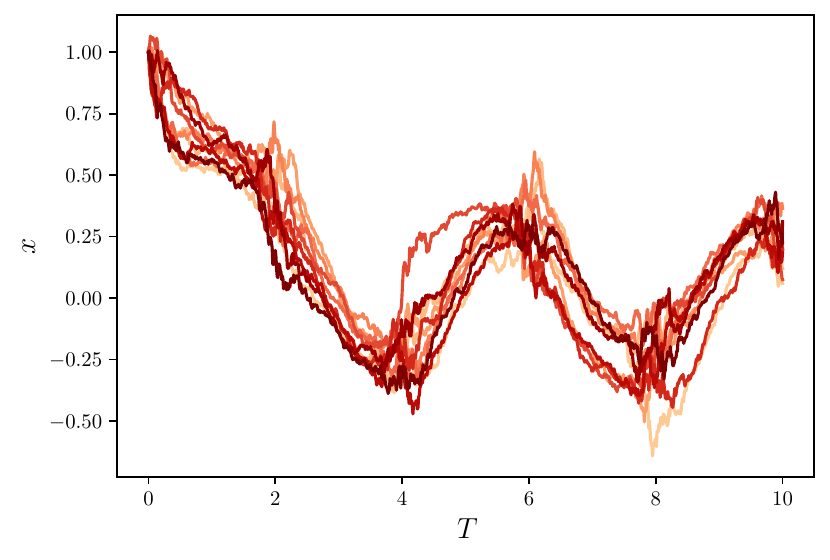}
  \caption{Sample trajectories of Example \ref{ex:OU2} with initial
    condition $x_0=1.0$, $\alpha(t)=\frac{1}{2}\sin(\frac{\pi}{2}t)$
    and $\beta(t)=\frac{1}{10}e^{\cos(\pi t)}$. Left: ground truth;
    Right: Simulation using the trained sFML model.}
\end{figure}
\begin{figure}[htbp]
  \centering
  \label{fig:OU_full_control1_ms}
  \includegraphics[width=.8\textwidth]{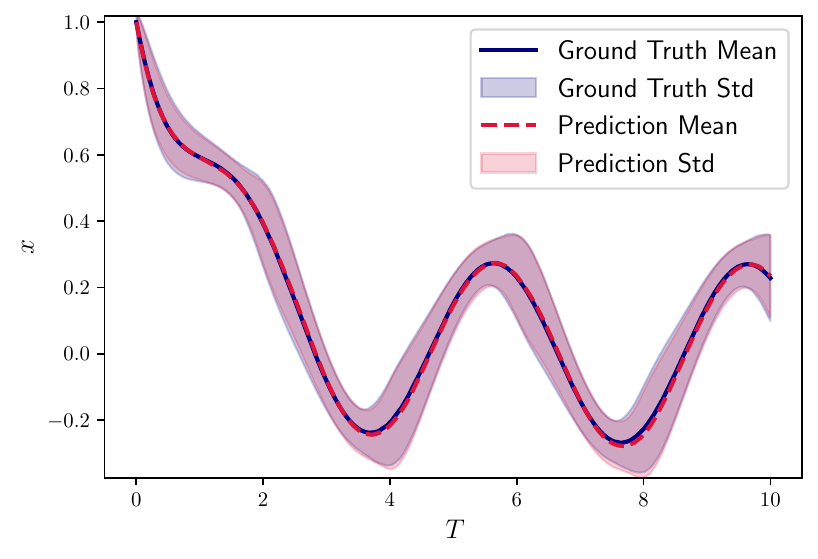}
  \caption{Mean and standard deviation (STD) of Example \ref{ex:OU2} with initial condition $x_0=1.0$, $\alpha(t)=\frac{1}{2}\sin(\frac{\pi}{2}t)$ and $\beta(t)=\frac{1}{10}e^{\cos(\pi t)}$.}
\end{figure}
\begin{figure}[htbp]
  \centering
  \label{fig:OU_full_control1_pdf}
  \includegraphics[width=.32\textwidth]{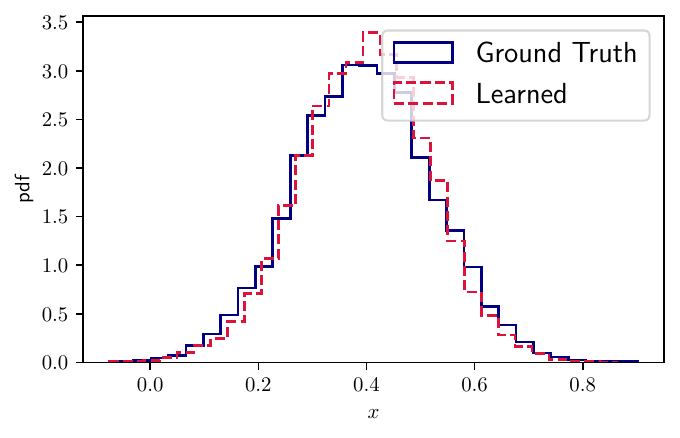}
  \includegraphics[width=.32\textwidth]{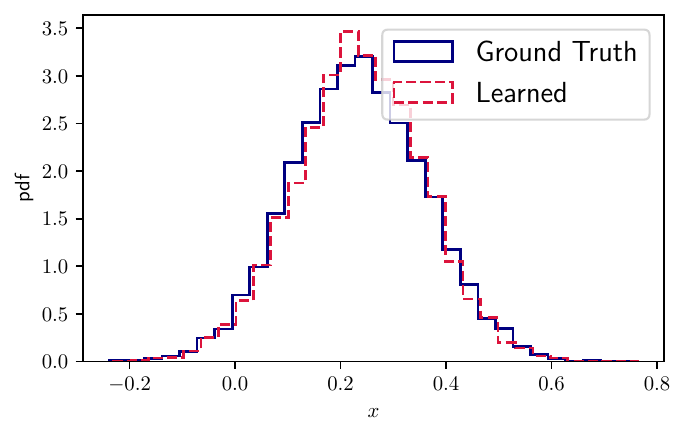}
  \includegraphics[width=.32\textwidth]{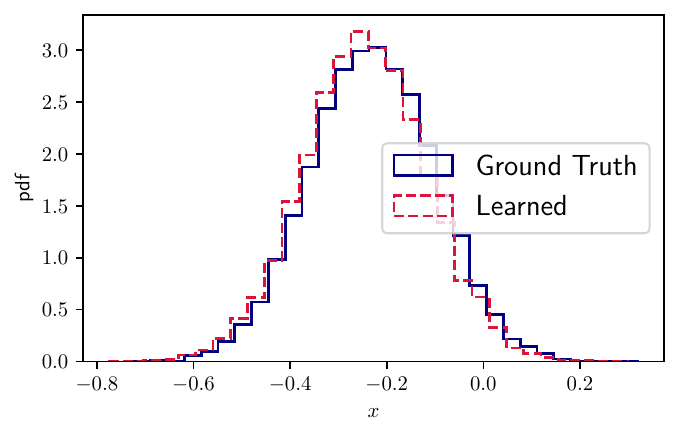}
  \caption{Comparasion of distribution of Example \ref{ex:OU2} at $T=2,6,8$ with initial condition $x_0=1.0$, $\alpha(t)=\frac{1}{2}\sin(\frac{\pi}{2}t)$ and $\beta(t)=\frac{1}{10}e^{\cos(\pi t)}$.}
\end{figure}

To examine the performance of the learned sFML model, we conduct a
simulation with an initial condition $\x_0=1.0$ and excitations
$\alpha(t)=\frac{1}{2}\sin(\frac{\pi}{2}t)$ and
$\beta(t)=\frac{1}{10}e^{\cos(\pi t)}$. (Note that the excitations are
not the Taylor \RV{polynomials} in the training data set.)  Some sample
solution trajectories are shown in Figure
\ref{fig:OU_full_control1_sample}. The mean and STD of the solution
are shown in Figure \ref{fig:OU_full_control1_ms}. And in
Figure \ref{fig:OU_full_control1_pdf}, we also show the comparison of
the probability distribution of the solution at $T=2,6,8$. We observe
good agreement between the sFML model prediction and the \RV{ground}
truth. 

\begin{comment}
We also test for initial condition $x_0=0.7$ and external inputs $\alpha(t)=\frac{1}{4}\sin(\pi t)$ and $\beta(t)=\frac{1}{5}+\frac{1}{10}\cos(\pi t)$. The sample trajectories under this setting are presented in Figure \ref{fig:OU_full_control2_sample}. The mean and STD are also validated, against the reference in Figure \ref{fig:OU_full_control2_ms}. Good agreement is observed in this case. 

\begin{figure}[htbp]
  \centering
  \label{fig:OU_full_control2_sample}
  \includegraphics[width=.48\textwidth]{./OU_stochas_control_re2/TruthSample0.pdf}
  \includegraphics[width=.48\textwidth]{./OU_stochas_control_re2/PredSample0.pdf}
  \caption{Sample trajectories of Example \ref{ex:OU2} with initial condition $x_0=0.7$, $\alpha(t)=\frac{1}{4}\sin(\pi t)$ and $\beta(t)=\frac{1}{5}+\frac{1}{10}\cos(\pi t)$. Left: ground truth; Right: Simulation using our trained model.}
\end{figure}

\begin{figure}[htbp]
  \centering
  \label{fig:OU_full_control2_ms}
  \includegraphics[width=.8\textwidth]{./OU_stochas_control_re2/Ex19BiStochsticOU_M1.pdf}
  \caption{Mean and standard deviation (STD) of Example \ref{ex:OU2} with initial condition $x_0=0.7$, $\alpha(t)=\frac{1}{4}\sin(\pi t)$ and $\beta(t)=\frac{1}{5}+\frac{1}{10}\cos(\pi t)$.}
\end{figure}
\end{comment}

\subsection{Nonlinear SDEs with Control}
\label{ex:nonlinear}
We now  consider a nonlinear system of SDEs, inspired by an exmple in Section 2.3.2 of \cite{weinan2011principles}: 
\RV{
\begin{equation}
    \left\{\begin{array}{l}
        d x_t = f(x_t, y_t, t)dt + \sigma_1 dW_1, \\
        d y_t =-\mu(y_t-x_t)dt +\sigma_2 dW_2,
    \end{array}\right.
\end{equation}
}
where $W_1$ and ${W}_2$ are independent Brownian motions, $\mu=1.0$,
$\sigma_1=0.2$, $\sigma_2=0.05$, and the function $f$ contains a control signal $u(t)$:
$$f(x, y, t)=-y^3+u(t), \qquad u(t)=\sin (\pi t)+\cos (\sqrt{2} \pi t).$$
To generate the training data, we simulate the system with $120,000$
sample paths over one time step $\Delta=0.01$ from initial conditions
uniformly in $I_\x=[-1.5,2.0] \times [-1.0,1.6]$ and under controls by
2nd-degree
Taylor polynomials with coeffficients sampled from $[-2,2] \times
[-8,8] \times [-15,15]$.
%The DNN used in our model is set with $3$ layers and $20$ nodes per layer. The model is trained for $300,000$ epochs.

\begin{figure}[htbp]
  \centering
  \label{fig:Nonlinear_sample}
  \includegraphics[width=.48\textwidth]{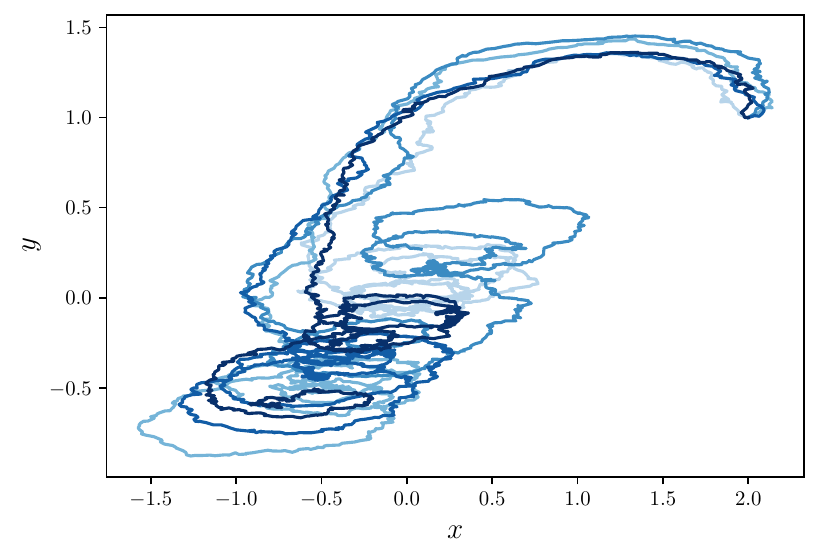}
  \includegraphics[width=.48\textwidth]{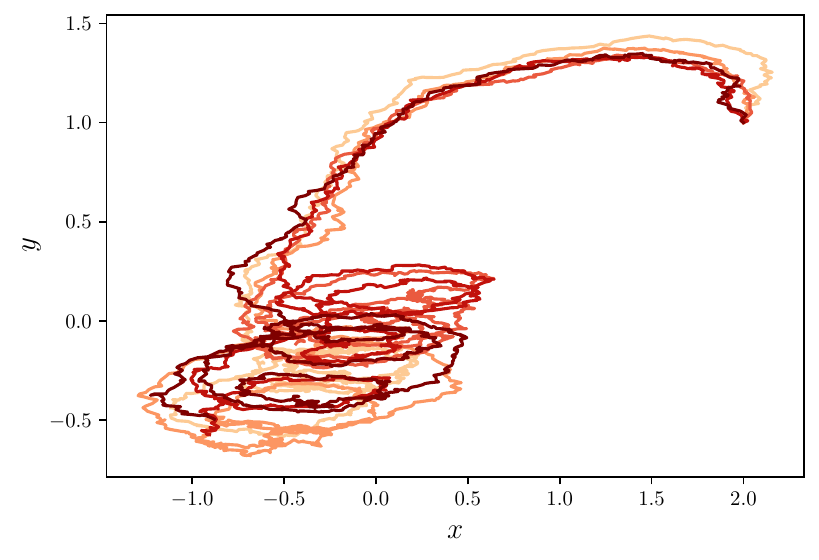}
  \caption{Sample phase portraits of Example \ref{ex:nonlinear} with
    initial condition $x_0=2.0$, $y_0=1.0$ for time up to $T=10$. Left: reference solution; Right: sFML model prediction.}
\end{figure}

\begin{figure}[htbp]
  \centering
  \label{fig:Nonlinear_ms}
  \includegraphics[width=.48\textwidth]{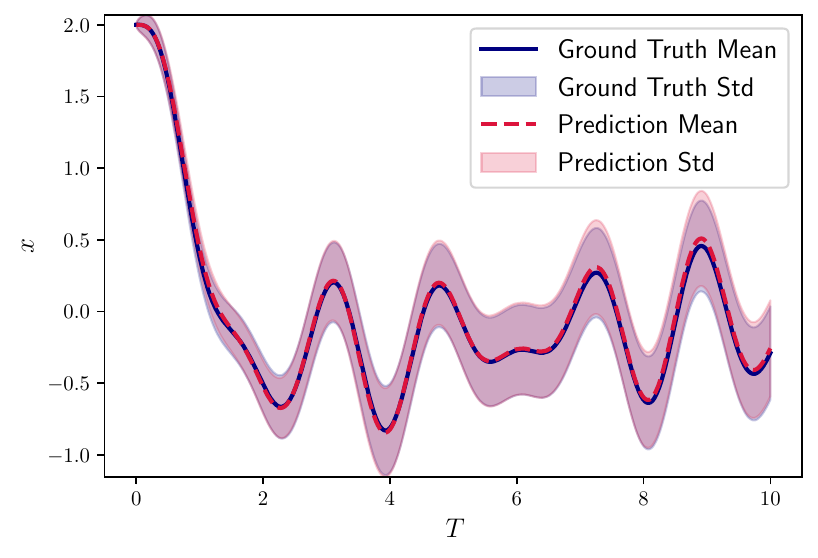}
  \includegraphics[width=.489\textwidth]{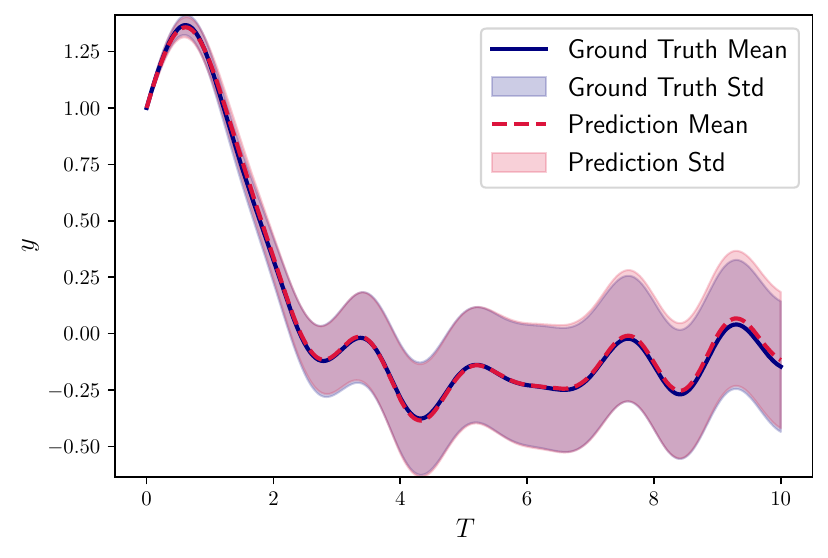}
  \caption{Mean and standard deviation (STD) of Example \ref{ex:nonlinear} with initial condition $x_0=2.0$, $y_0=1.0$. Left: $x$; Right: $y$.}
\end{figure}

\begin{figure}[htbp]
  \centering
  \label{fig:Nonlinear_pdf}
  \includegraphics[width=.24\textwidth]{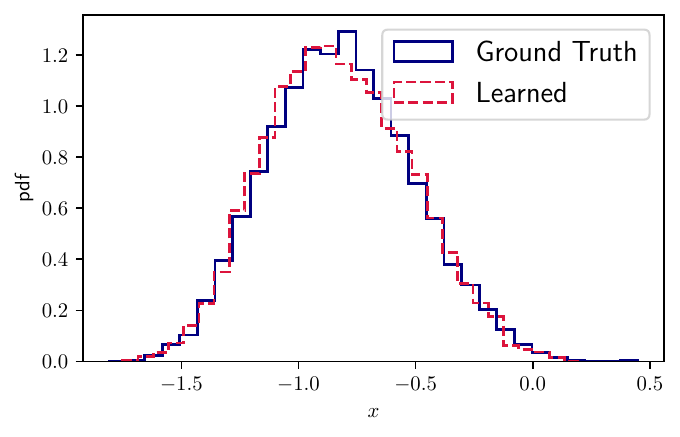}
  \includegraphics[width=.24\textwidth]{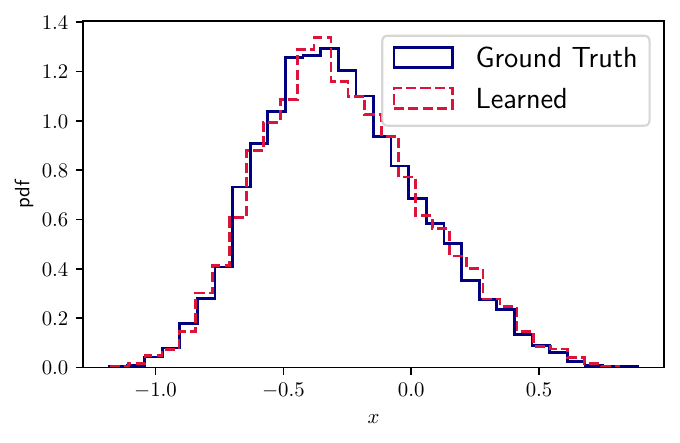}
  \includegraphics[width=.24\textwidth]{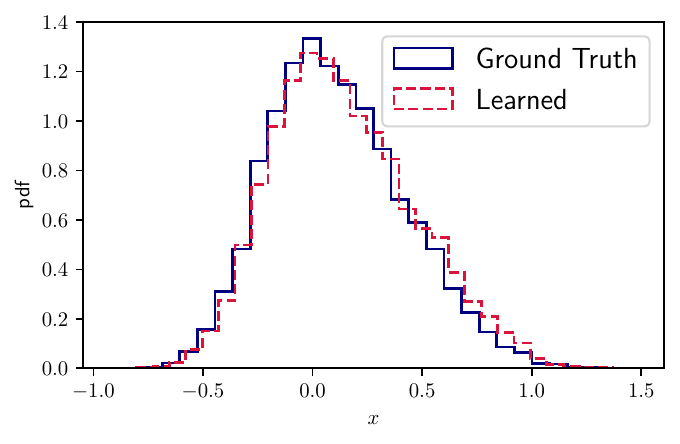}
  \includegraphics[width=.24\textwidth]{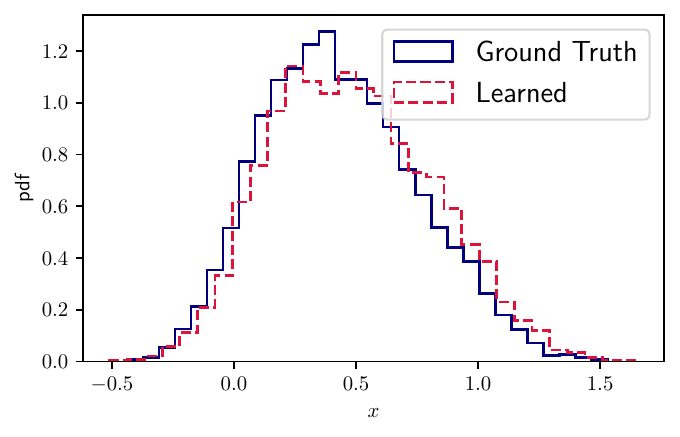}
  \includegraphics[width=.24\textwidth]{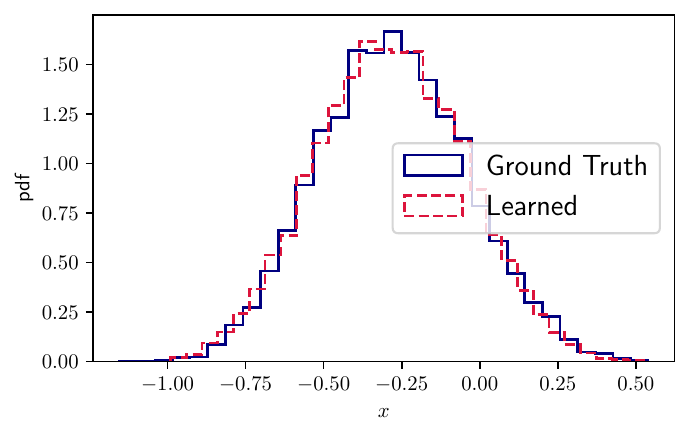}
  \includegraphics[width=.24\textwidth]{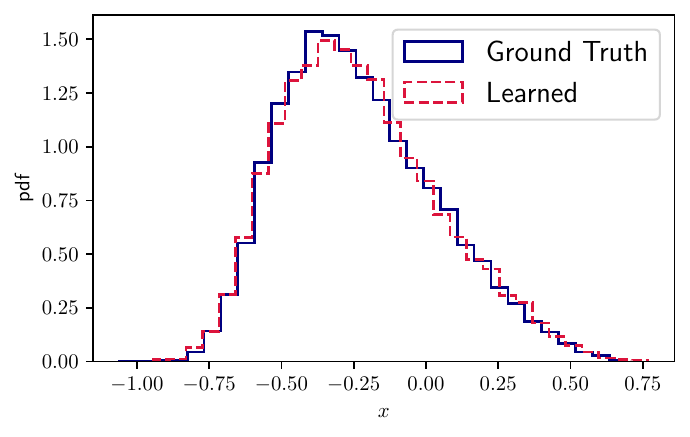}
  \includegraphics[width=.24\textwidth]{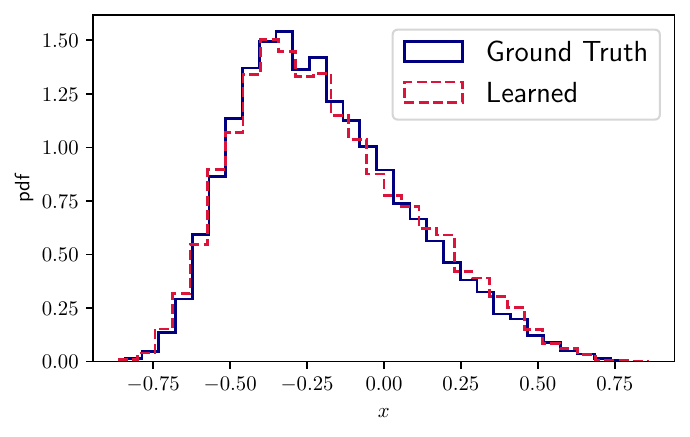}
  \includegraphics[width=.24\textwidth]{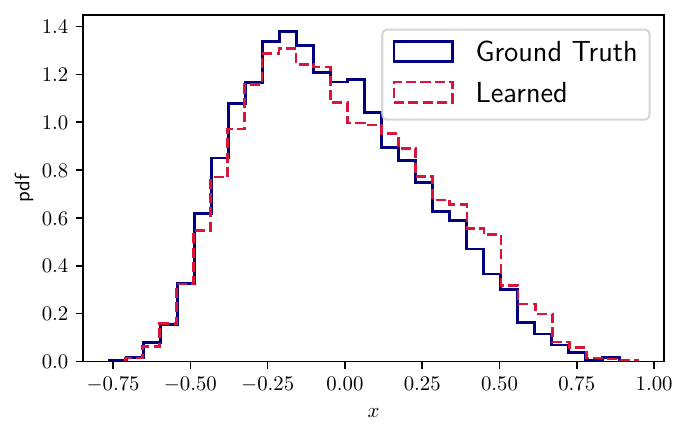}
  \caption{Comparasion of probability distributions of Example
    \ref{ex:nonlinear} at $T=4,6,7,9$ with an initial condition
    $x_0=2.0$, $y_0=1.0$. Top row: $x$; Bottom row: $y$.} 
\end{figure}

For the learned sFML model, we conduct system predictions with an
initial condition $x_0=2.0$ and $y_0=1.0$. In Figure
\ref{fig:Nonlinear_sample}, we plot a few sample phase portraits from
ground truth (left), as well as from the sFML model prediction
(right). They appear to be visually in agreement. The mean and
standard deviation of the system prediction by the sFML model are
shown in Figure \ref{fig:Nonlinear_ms}, along with those of the true
solution. In Figure \ref{fig:Nonlinear_pdf}, we also show the
comparison of reference and learned density functions of the test
trajectory at time $T=4,6,7,9$. We observe that the sFML model
exhibits good accuracy in these predictions.

\subsection{Stochastic Predator-Prey Model}
\label{ex:pre}
We then consider a stochastic Lotka-Volterra system with a time-dependent excitation $u(t)$:
\RV{
\begin{equation}
    \left\{\begin{array}{l}
        dx_t = \left [ x_t -x_t y_t +u(t) \right ]dt + \sigma_1 x_t dW_1, \\
        dy_t = (-y_t + x_t y_t)dt +\sigma_2 y_t dW_2,
    \end{array}\right.
\end{equation}
}
where ${W}_1$ and ${W}_2$ are independent Brownian motions, and $\sigma_1=\sigma_2=0.05$.
The training data are generated by simulating $120,000$ \RV{solution} samples for one step $\Delta=0.01$, from initial conditions in  $I_\x=[0.1,0.35] \times [0.2,5.5]$ and under \RV{excitations} of 2nd-degree Taylor polynomials whose coefficients are from $[0.01,4.2] \times [-1.5,1.5] \times [-0.7,0.7]$. 

\begin{figure}[htbp]
  \centering
  \label{fig:Pred_sample}
  \includegraphics[width=.48\textwidth]{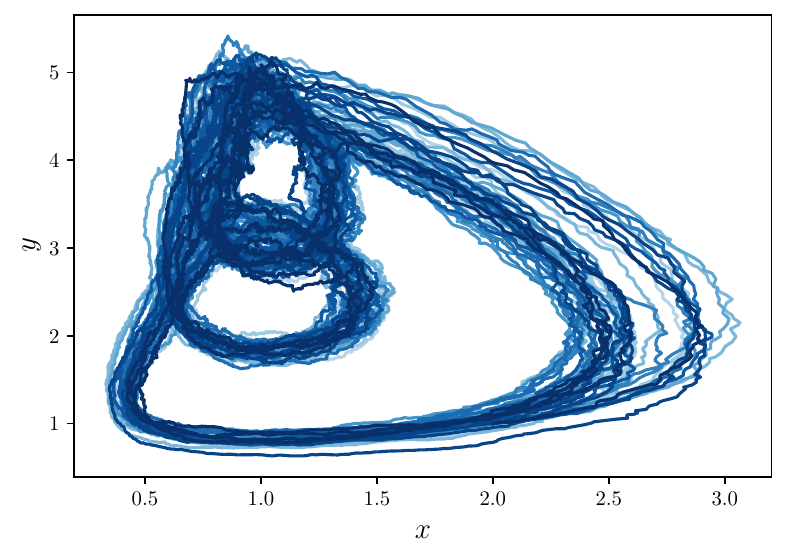}
  \includegraphics[width=.48\textwidth]{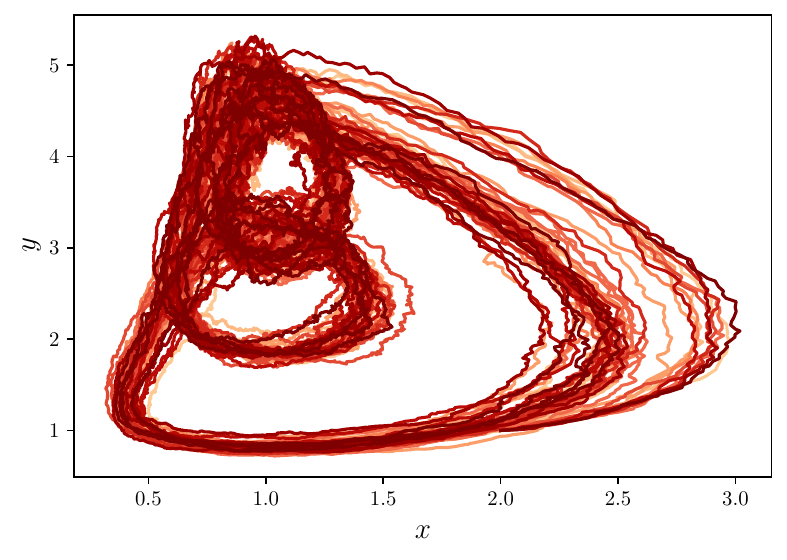}
  \caption{Phase portrait samples of the Lotka-Volterra system \ref{ex:pre} with an initial condition $x_0=2.0$, $y_0=1.0$ and $u(t)=\sin(\frac{t}{3})+\cos(t)+2$.}
\end{figure}

\begin{figure}[htbp]
  \centering
  \label{fig:Pred_ms}
  \includegraphics[width=.98\textwidth]{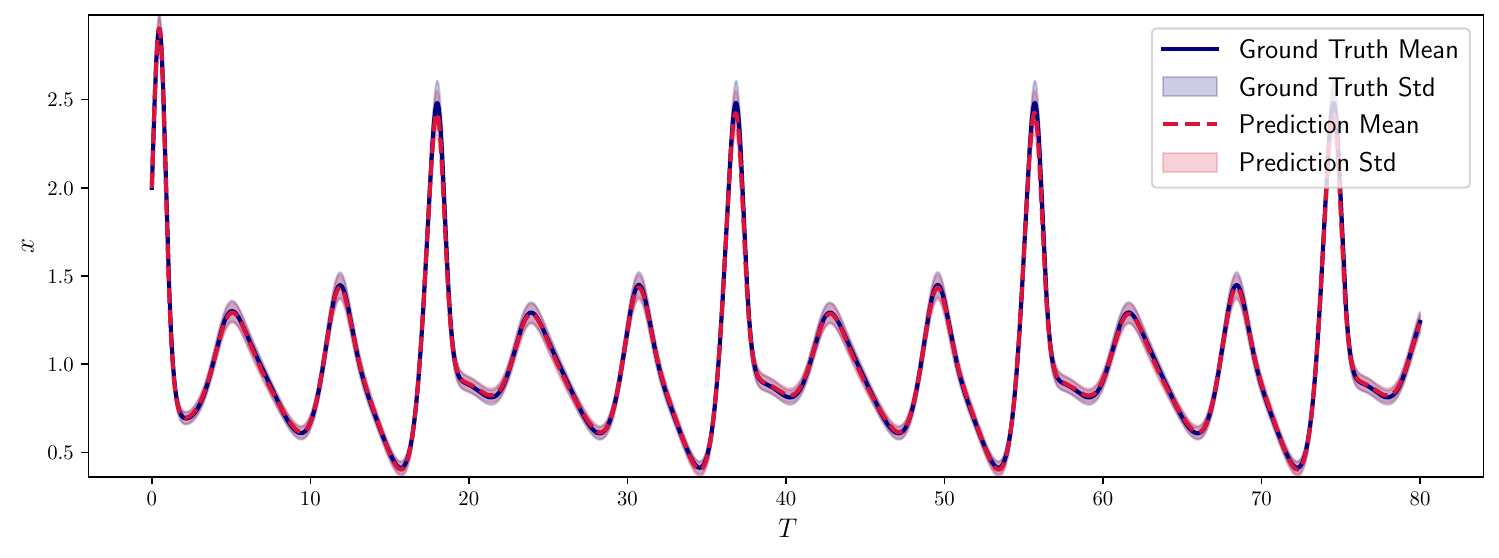}
  \includegraphics[width=.98\textwidth]{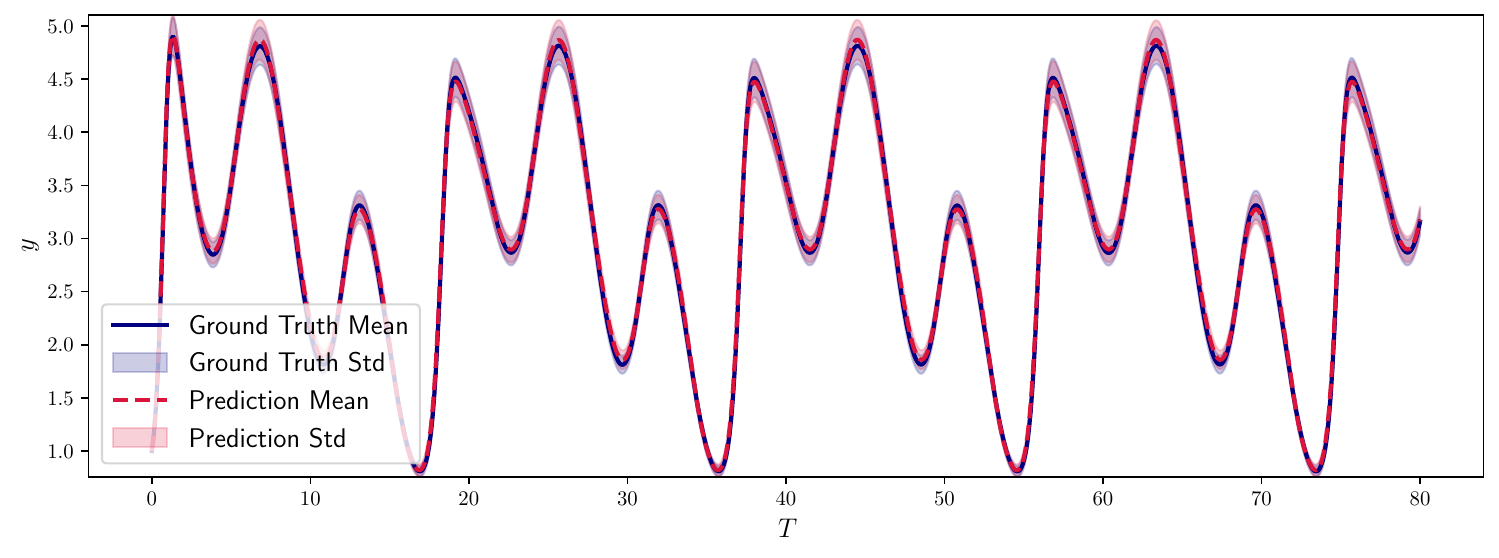}
  \caption{Mean and standard deviation (STD) of Example \ref{ex:pre} with initial condition $x_0=2.0$, $y_0=1.0$ and $u(t)=\sin(\frac{t}{3})+\cos(t)+2$. Upper: $x$; Lower: $y$.}
\end{figure}

Once we have the trained model, we conduct system prediction with an initial condition $x_0=2.0$, $y_0=1.0$ and exitation $u(t)=\sin(\frac{t}{3})+\cos(t)+2$. We conduct relatively long-term prediction for time up to $T=80$. (Note that the training data are of lenght $0.01$.) In  Figure \ref{fig:Pred_sample}, we plot a few sample of the phase portrait of the system. Good visual agreement between the sFML prediction and the ground truth can be observed. To examine the accuracy more closely, we present the mean and standard deviation of the system in Figure \ref{fig:Pred_ms}. We observe good predictive accuracy of the sFML model for up to $T=80$.

\subsection{Stochastic Resonance}
\label{ex:StochasticRes}
\RV{In this section}, we consider the following SDE  with a double-well potential and excitation,
\begin{equation} \label{equ:SR}
    dx_t = \left[x_t-x_t^3 + u(t) \right] dt+\sigma dW_t,
\end{equation}
where $\sigma=0.25$ is a parameter, and $u(t)$ is the excitation. When
$V=0$, there is no excitation to the system. The solution would
exhibit random transition between two metastable states $x=-1$ and
$x=1$. The transition probability depends on the parameters
$\sigma$. When $V\neq 0$, an excitation 
is exerted to the system. If the excitation is periodic, under the
right circumstance the random \RV{transition} between the two metastable
states becomes \RV{synchronized} with the \RV{periodicity} of the \RV{excitation},
resulting in the so-called stochastic resonance, cf.,
\cite{benzi1981mechanism,benzi1982stochastic,benzi1983theory}. 
%
%The autonomous double well system (i.e. without external force, $V=0$ here) switches between two metastable states $x=1$ and $-1$ randomly. In the recent work \cite{chen2023learning,xu2023learning}, sFM models are used to recover the trajectories of this system. If the system is applied with an external weak periodic force as in \eqref{equ:SR}, the hopping between two states can become synchronized with the periodicity of the force. This only happens when the noise level $\sigma$ is taken to be a proper value. The mechanism of this phenomenon is extensively studied by \cite{benzi1981mechanism,benzi1982stochastic,benzi1983theory} to explain the periodically recurrent ice ages.

Here, we demonstrate that the proposed sFML method can accurately
model and predict the long-term system behavior using only very short
burst of measurement data.
Our data are $30,000$
trajectories of one step ($\Delta=0.01$) length, with initial
conditions sampled from $I_\x=[-1.6,1.6]$ and under piecewise constant
exictations sampled from $[-0.13,0.13]$. 
%In our present work, we are trying to learn this system. The external input $\alpha(t):=V\cos(\omega t)$ is locally parameterized by piecewise constant, i.e. polynomials of degree $0$. Thus we have $\Gam_n \in \mathbb{R}$.  More specifically, we generate $30,000$ training data with initial conditions sampled from and. Due to the long simulation period, we set $\Delta=0.1$ in the training data generation. The DNN used in our model is set with $3$ layers and $20$ nodes per layer. We train the model with $200,000$ epochs.

\begin{figure}[htbp]
  \centering
  \label{fig:SR_sample}
  \includegraphics[width=.99\textwidth]{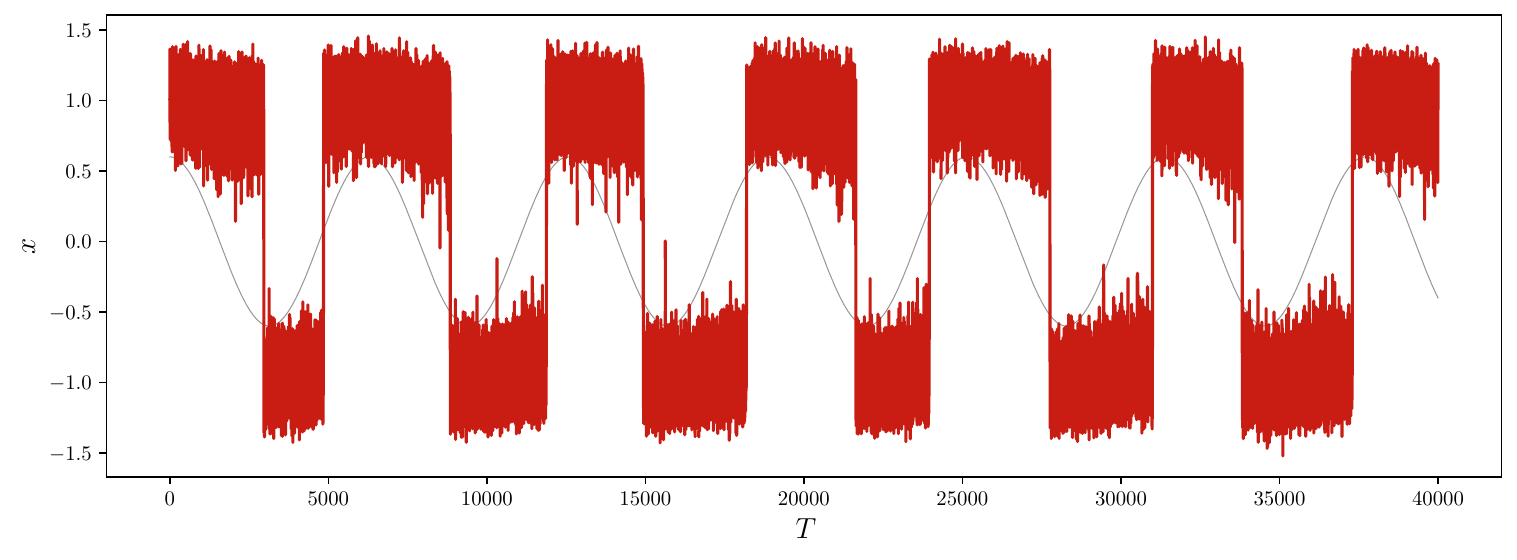}
  \includegraphics[width=.99\textwidth]{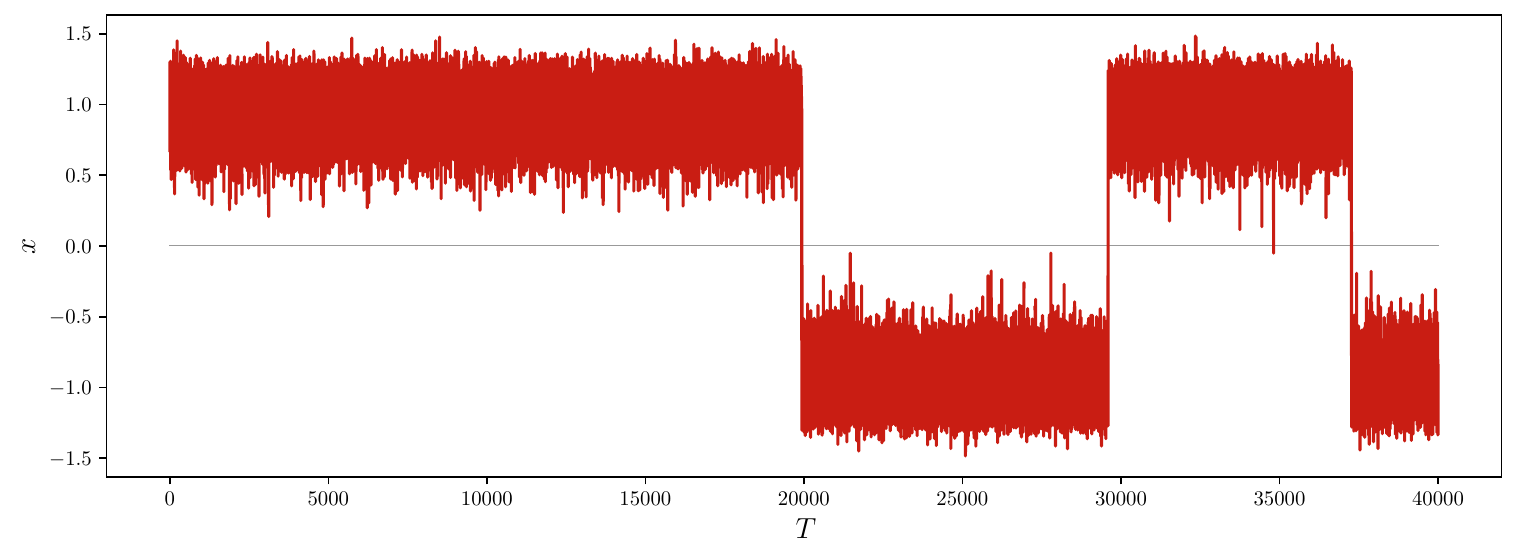}
  \caption{Sample trajectories of Example \ref{ex:StochasticRes} with
    an initial condition $x_0=1.0$. Top: With the periodic excitation
    $u(t)=V\cos(\omega t)$, $V=0.12$ and $\omega=0.001$. The system
    exhibits stochastic resonance. Bottom: No excitation case with
    $u(t)=0$. The system exhibits random transitions with very small
    probability $O(10^{-5})$. }
\end{figure}

\begin{figure}[htbp]
  \centering
  \label{fig:SR_pdf}
  \includegraphics[width=.24\textwidth]{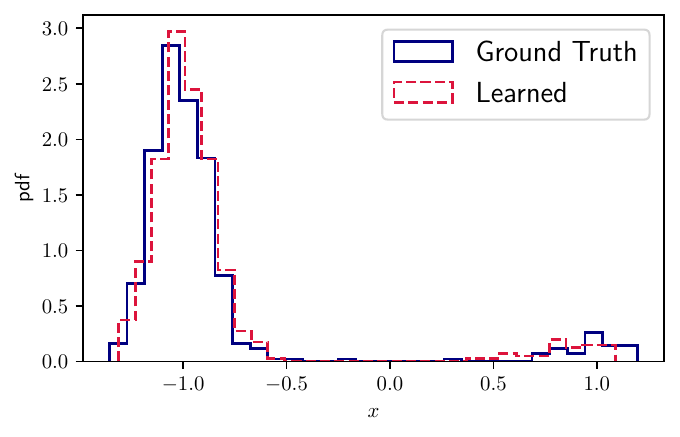}
  \includegraphics[width=.24\textwidth]{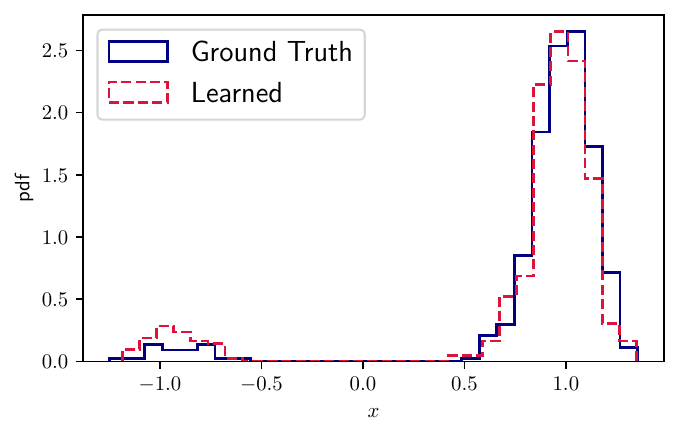}
  \includegraphics[width=.24\textwidth]{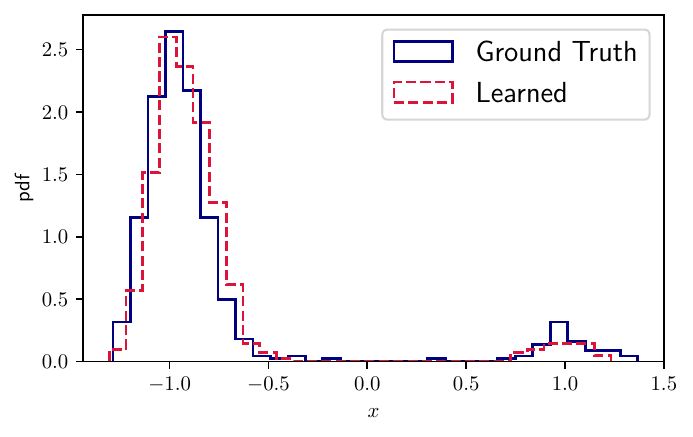}
  \includegraphics[width=.24\textwidth]{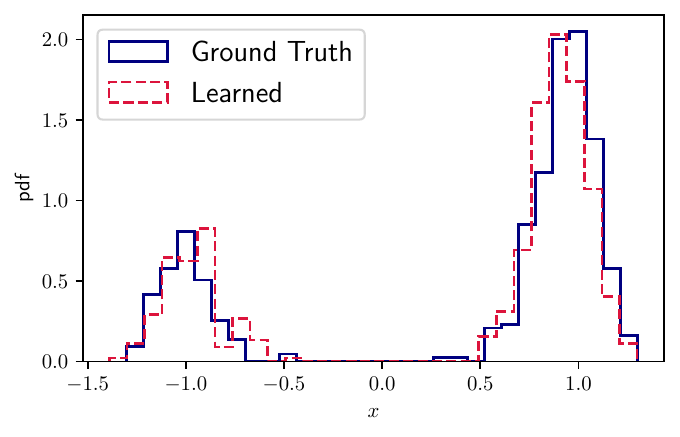}
  \caption{\RV{Comparasion of probability distributions of Example
    \ref{ex:StochasticRes} at $T=10,000,20,000,30,000,40,000$ with the periodic excitation
    $u(t)=V\cos(\omega t)$, $V=0.12$ and $\omega=0.001$}.} 
\end{figure}

Once the sFML model is trained, we conduct system prediction under
various excitations. In particular, we choose $u(t)=V\cos(\omega t)$,
with $V=0.12$ and $\omega=0.001$. These parameters are chosen
according to \cite{benzi1981mechanism}, to ensure the occurrence of
stochastic resonance. An \RV{exceptionally} long-term system prediction is
conducted by the sFML model, for time up to $T=40,000$. The result is
shown in the top of Figure \ref{fig:SR_sample}, where we also plotted
the (rescaled) periodic excitation in light grey line in the
background. We can clearly observe the \RV{synchronization} between the
random transition and the periodic excitation --- the stochastic
resonance. 
For reference, we also conduct the sFML system prediction with $V=0$,
i.e., no excitation. The solution, shown in the bottom of  Figure
\ref{fig:SR_sample}, exhibits the expected random transition between
the two metastable states. We shall \RV{emphasize} that in this case the
transition probability is very small, $O(10^{-6})$. The learned sFML
model is capable of capturing such a small probability event. We shall
remark again that the training data are pairwise data separated by one
time step. Thus, none of the (long-term) system behaviors can be
observed in the training data. \RV{In Figure \ref{fig:SR_pdf}, we also show the
comparison of reference and learned density functions of the test
trajectory at time $T=10,000,20,000,30,000,40,000$. We observe that the sFML model
exhibits good accuracy in these predictions.}
%Using the trained model, we generate the sample trajectories in several scenarios. We first take the parameters and $\sigma=0.25$ in \eqref{equ:SR} (as in \cite{benzi1981mechanism}) and predict the system up to $T=40,000$.
%.A sample trajectory is presented in Figure \ref{fig:SR_sample}, in which the trajectory is plotted with red color, while the (scaled) periodic force is plotted with grey color. The simulation result is shown to be stable in an extremely long time run, while the switch of metastates is observed to be synchronized with the periodicity of external force. To compare with this scenario, we also predict with $V=0$, i.e. without external inputs, up to $T=40,000$. One trajectory under this setting is presented in Figure \ref{fig:SR_sample}. In this case, as we expected, the trajectory switches between metastable states randomly. 

\RV{
\subsection{A Gene Expression Model}
\label{chem}
We consider a gene expression model in the form of biochemical
reaction network \cite{voliotis2016stochastic,bowsher2013fidelity},
described as Example 2 in Section \ref{sec:setup},
\begin{equation}
    \begin{split}
        \emptyset     & \xrightarrow{k(t)} M \\
        M & \xrightarrow{k_\text{s}} M+P \\
        M & \xrightarrow{k_{\text{dm}}} \emptyset \\
        P & \xrightarrow{k_{\text{dp}}} \emptyset,
    \end{split}
\end{equation}
where there are 2 species: $M$ represents mRNA and $P$ represents
protein. In this two-stage reaction network model, mRNAs are
transcribed in a time-dependent rate $k(t)$ and proteins are
translated at a rate of $k_\text{s}$ off each mRNA
molecule. Meanwhile, mRNAs and proteins degrade at rate
$k_{\text{dm}}$ and $k_{\text{dp}}$. The dynamical behavior of state
variables $(M,P)$ is usually modeled as a stochastic jump process due
to the stochasticity caused by low number of molecules. Stochastic
Simulation Algorithm (SSA, c.f. \cite{gillespie1977exact}) is an exact
Monte Carlo method designed for producing realizations of such
processes. Though the transitional probability of these processes is
non-Gaussian and nearly impossible to be explicitly written, we
demonstrate in this example that the proposed sFML method is capable
of learning and predicting such systems. 

We simulate the system using Modified Next Reaction method
\cite{anderson2007modified}, which is a variant of the original
SSA. The training data are generated via simulating the system up to a
short time $\Delta = 0.1$ with randomly sampled initial conditions for
$M\leq 10$ and $P\leq 400$. Taylor polynomials of 
degree $2$ are used to conduct local approximation of the excitation signal
$k(t)$ within the $\Delta$ time interval.
%The initial conditions $(M_0,P_0)$ are uniformly sampled from
%$[-5,10]\times[-200,400]$ and then projected onto the half space
%$[0,\infty)^2$ which leads to higher weights on the domain
%boundary. This special treatment aims to enhance the approximation
%accuracy for domain boundary points, where rare events usually
%happen.
The 3 parameters for $\Gam_0$ are sampled uniformly from
$[0,40]\times[-5.3,5.3]\times [-0.7,0.7]$. For other parameters, we fix
$\Delta=0.1$, $k_\text{s}=500$, $k_{\text{dm}}=20$,
$k_{\text{dp}}=5$. (Note that these parameters are only used to
generate the training data. They are not included in the FML model.)

For the learned sFML model, we conduct a system prediction with an initial
condition $(M_0,P_0)=(1,1)$ and $k(t)=20+20\sin(\frac{\pi x}{12})$ up
to time $T=240$. To test the learning accuracy, we fix
$(\x_0,\Gam_0)$, generate $10,000$ samples of $\z$ for
$\G_\Delta(\x_0,\z;\Gam_0)$ to compare the conditional
distribution. In Figure \ref{fig:chem_condpdf}, we show the 2D
histogram for 2 scenarios: $\x_0=(2,133)$, $\Gam_0=(30, -4.535,
-0.335)$ and $\x_0=(5, 287)$, $\Gam_0=(39.319,  1.355, -0.664)$. The
sFML model achieves high accuracy in this non-Gaussian transitional
probability distribution case. 

\begin{figure}[htbp]
  \centering
  \label{fig:chem_condpdf}
  \includegraphics[width=.24\textwidth]{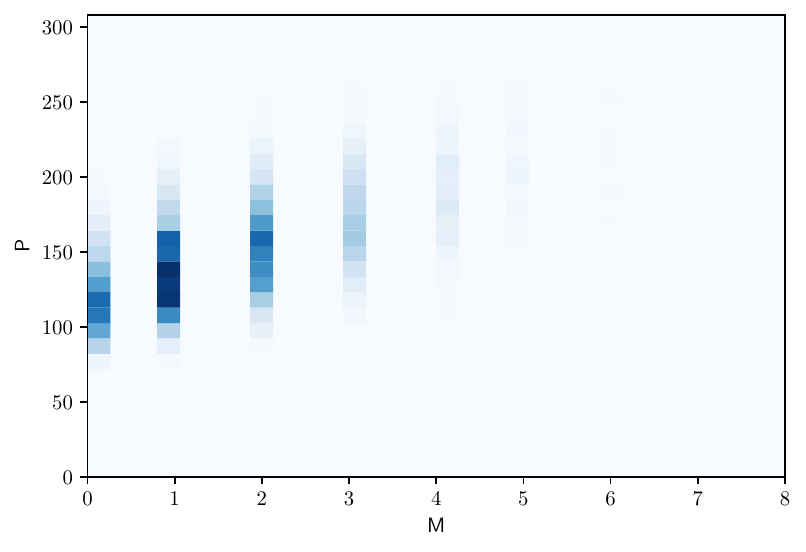}
  \hspace{-2mm}
  \includegraphics[width=.24\textwidth]{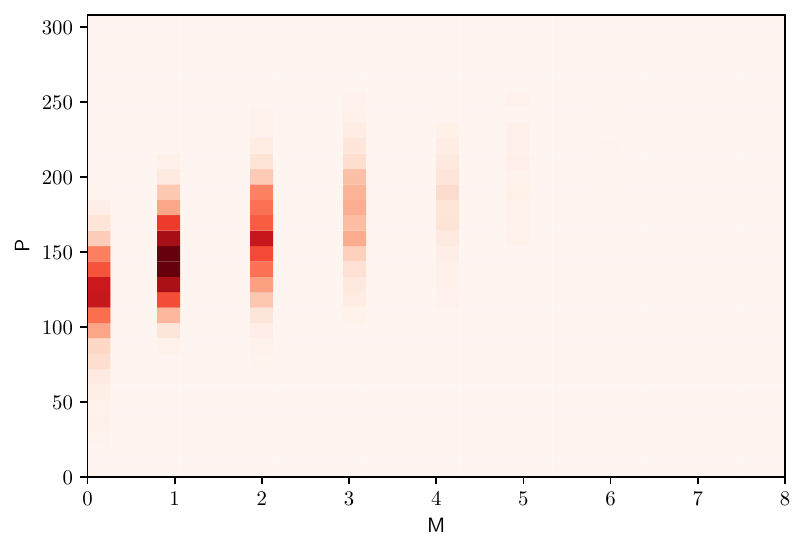}
  \hspace{1mm}
  \includegraphics[width=.24\textwidth]{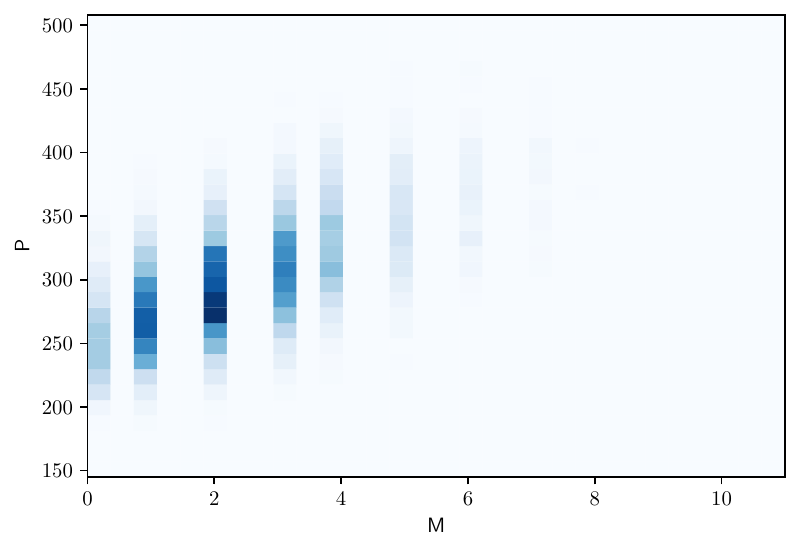}
  \hspace{-2mm}
  \includegraphics[width=.24\textwidth]{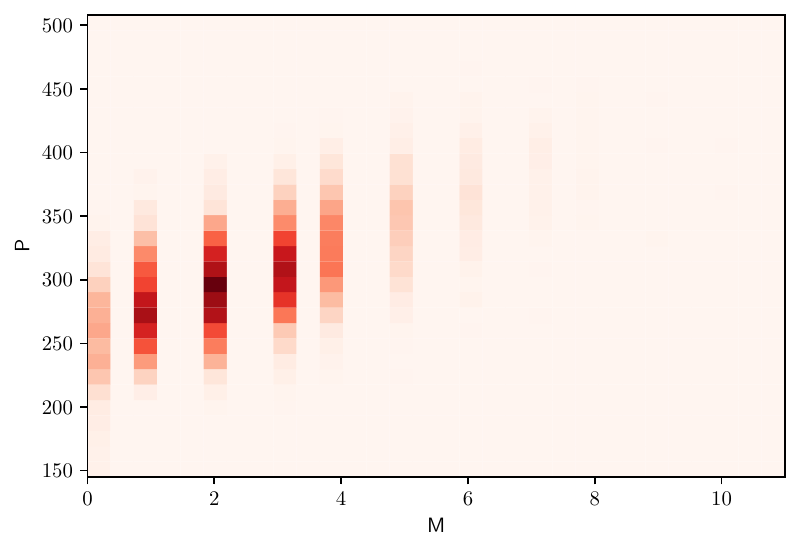}
  \caption{\RV{2D histogram of (non-Gaussian) conditional distribution $\G_\Delta(\x_0,\z;\Gam_0)$ with fixed $(\x_0,\Gam_0)$ of Example \ref{chem}. Left 2 figures: reference and learned distribution for $\x_0=(2,133)$ and $\Gam_0=(30, -4.535, -0.335)$; Right 2 figures: reference and learned distribution for $\x_0=(5, 287)$ and $\Gam_0=(39.319,  1.355, -0.664)$.}} 
\end{figure}

\begin{figure}[htbp]
  \centering
  \label{fig:chem_sample}
  \includegraphics[width=.49\textwidth]{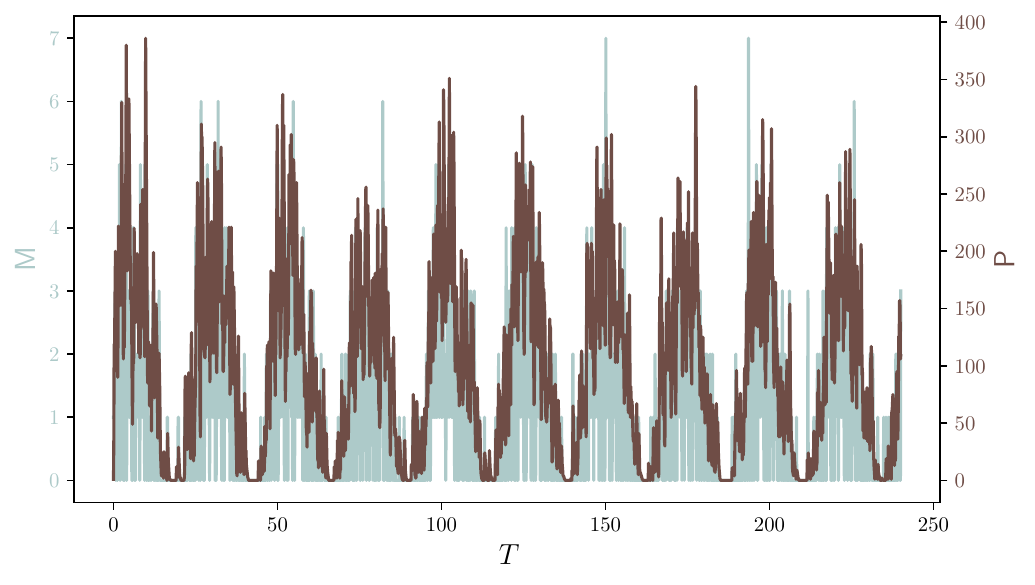}
  \includegraphics[width=.49\textwidth]{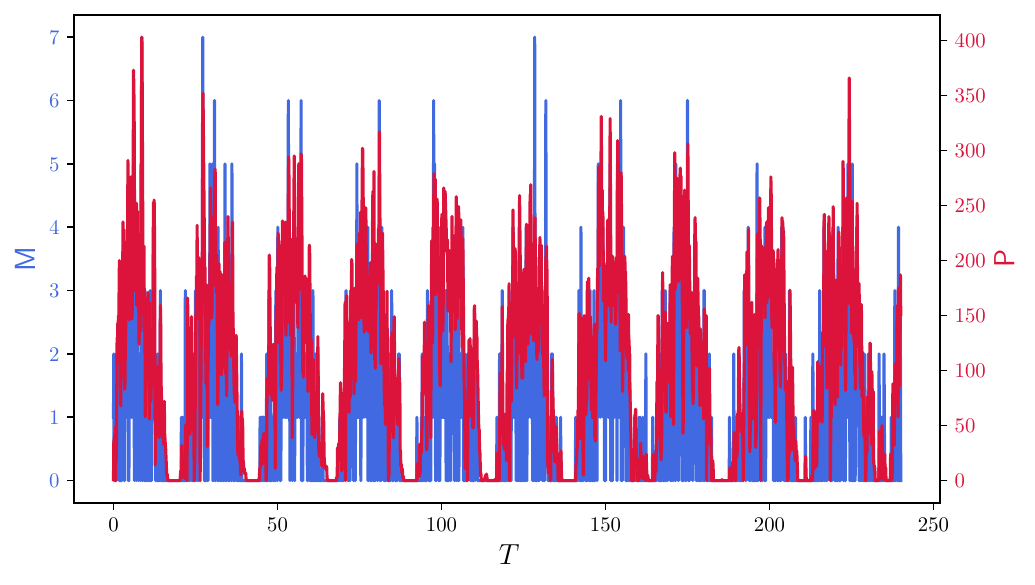}
  \caption{\RV{Sample trajectory for Example \ref{chem} with initial condition $(M_0,P_0)=(1,1)$ and $k(t)=20+20\sin(\frac{\pi x}{12})$. Left: a reference sample trajectory; Right: a learned sample trajectory.}} 
\end{figure}

\begin{figure}[htbp]
  \centering
  \label{fig:chem_meanstd}
  \includegraphics[width=.49\textwidth]{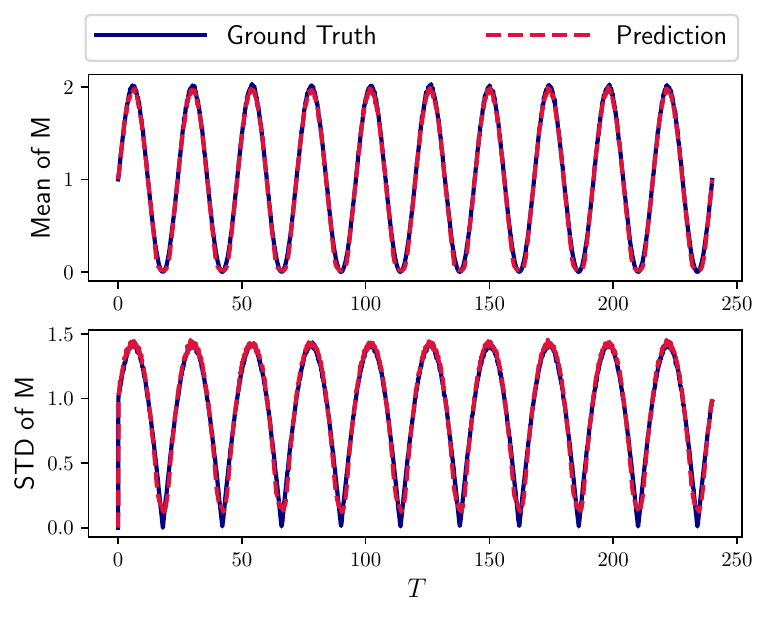}
  \includegraphics[width=.49\textwidth]{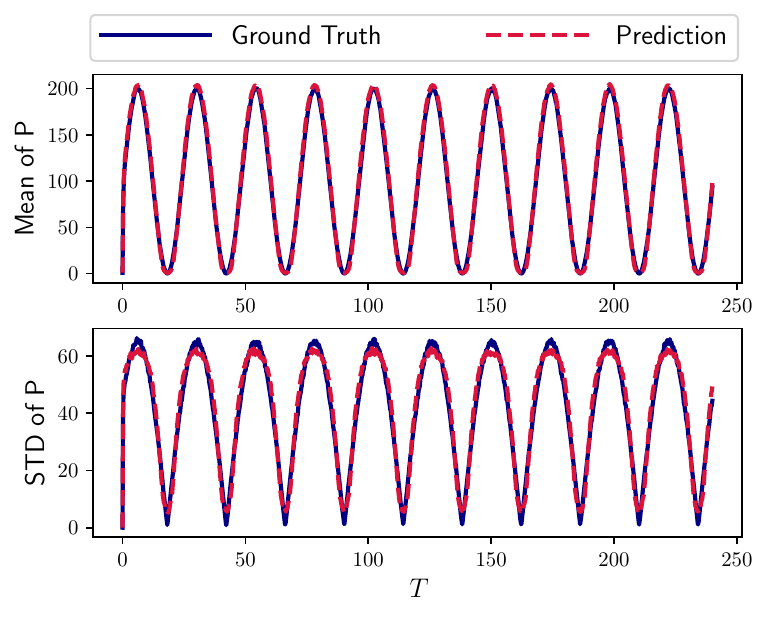}
  \caption{\RV{Mean and standard deviation (STD) for Example \ref{chem} with initial condition $(M_0,P_0)=(1,1)$ and $k(t)=20+20\sin(\frac{\pi x}{12})$. Left: $M$; Right: $P$.}} 
\end{figure}

\begin{figure}[htbp]
  \centering
  \label{fig:chem_pdf}
  \includegraphics[width=.24\textwidth]{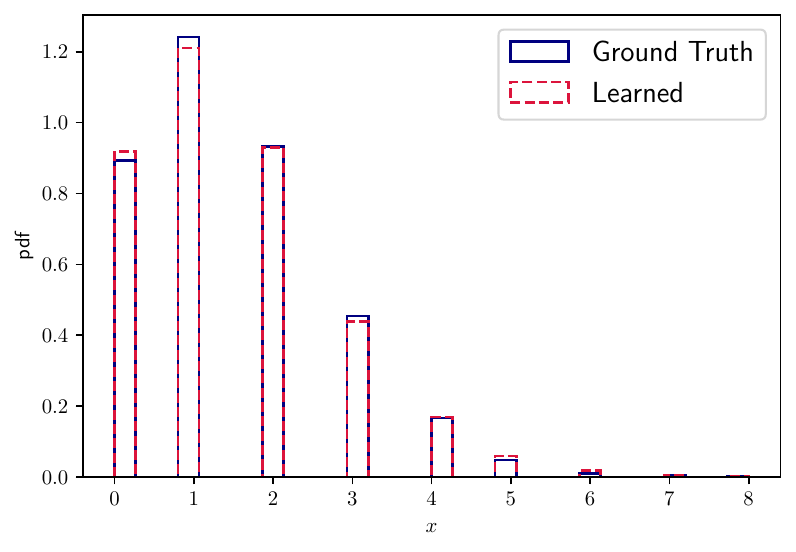}
  \includegraphics[width=.24\textwidth]{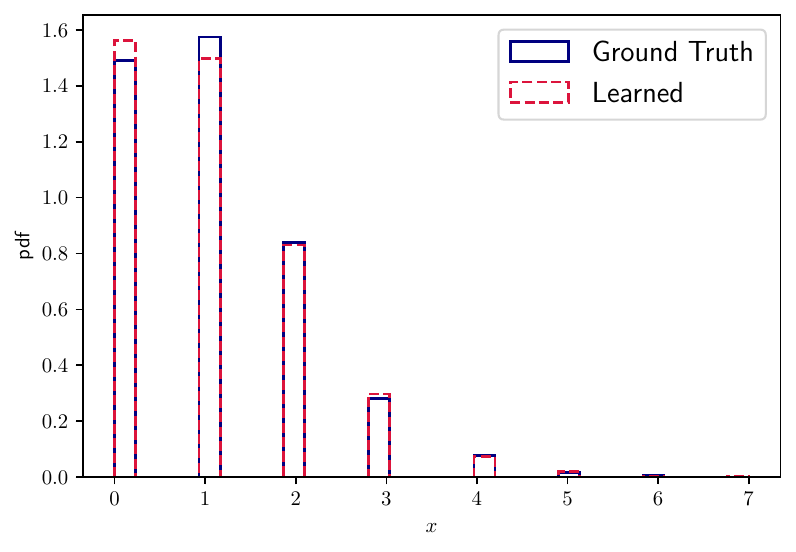}
  \includegraphics[width=.24\textwidth]{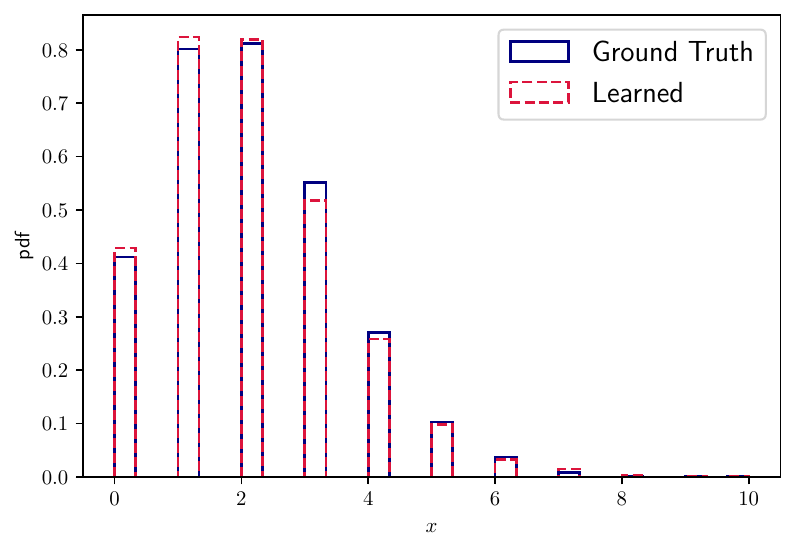}
  \includegraphics[width=.24\textwidth]{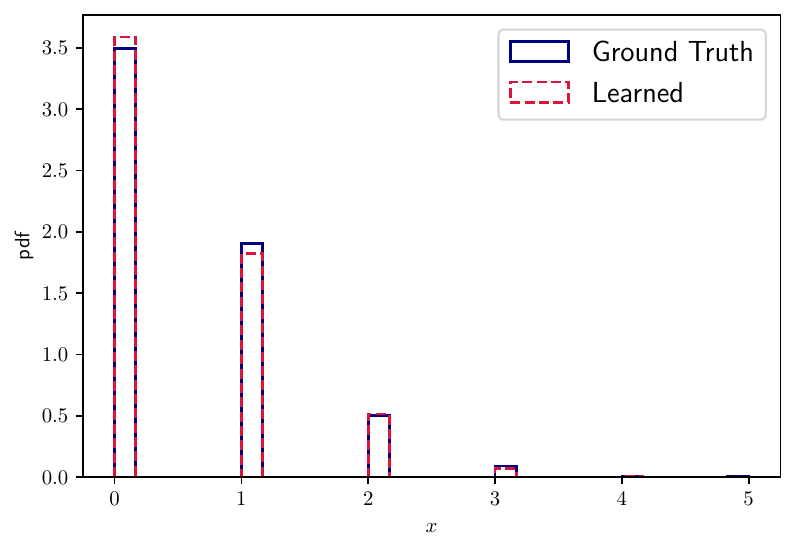}
  \includegraphics[width=.24\textwidth]{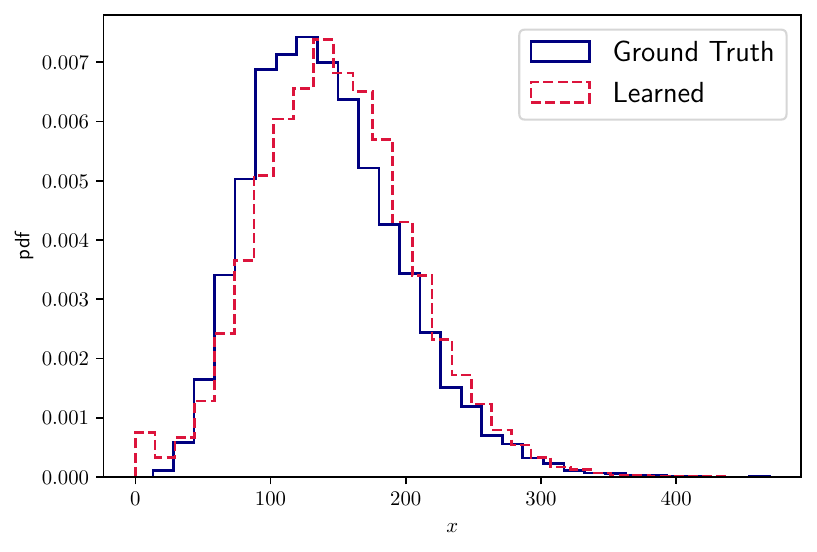}
  \includegraphics[width=.24\textwidth]{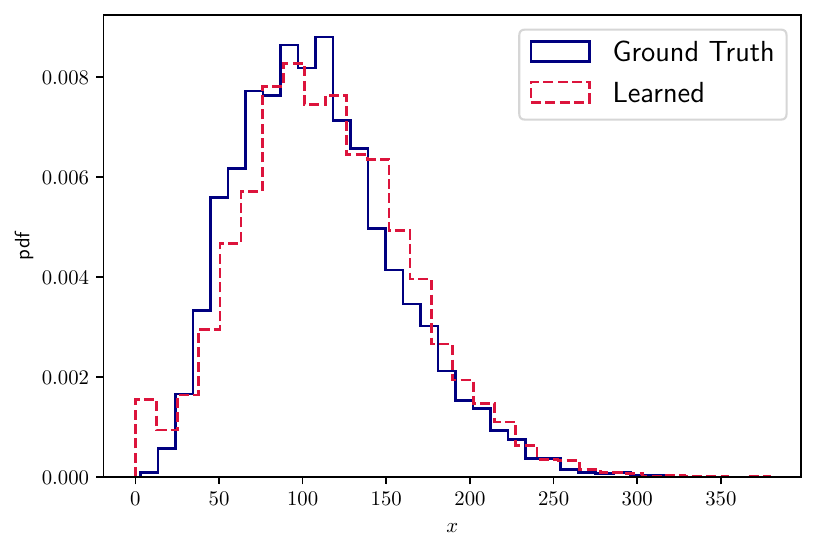}
  \includegraphics[width=.24\textwidth]{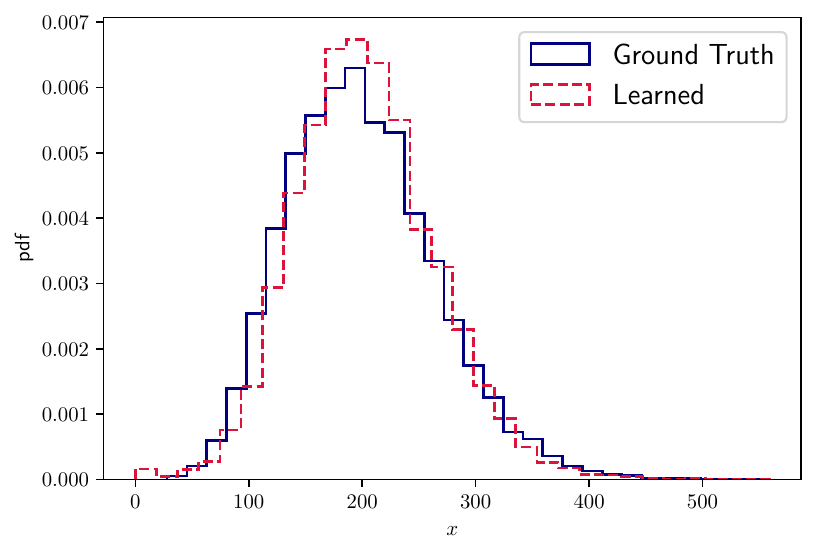}
  \includegraphics[width=.24\textwidth]{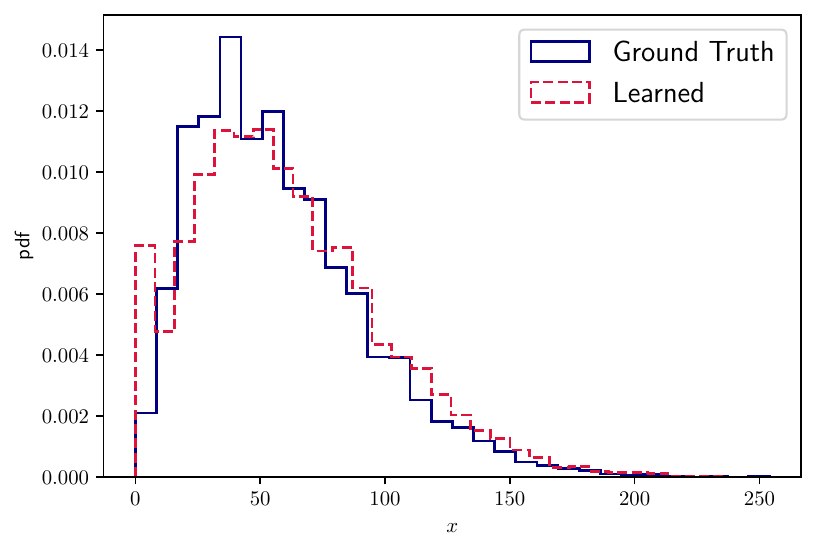}
  \caption{Comparasion of probability distributions of Example \ref{chem} at $T=50,60,78,110$ with initial condition $(M_0,P_0)=(1,1)$ and $k(t)=20+20\sin(\frac{\pi x}{12})$. Top row: $M$; Bottom row: $P$.} 
\end{figure}

In Figure \ref{fig:chem_sample}, we show one trajectory produced by
SSA (left) and the learned sFML model (right). These trajectories
reveal similar patterns. To further validate the long time accuracy,
we compute the mean and standard deviation with $10,000$ trajectories
and show the results in Figure \ref{fig:chem_meanstd}. Moreover, the
probability distributions at time $T=50$,$60$,$78$, $110$ are shown in Figure
\ref{fig:chem_pdf}. We observe good agreement between the sFML model
predictions and the ground truth.  

}

\RV{
\subsection{A Stochastic Heat Equation with Source}
In this section, we consider learning a one-dimensional stochastic heat equation driven by space-time noise
\begin{align}\label{sheat}
 \partial_t u    = \varepsilon\Delta u + f(x,t) + \sigma
   \xi(x,t), \qquad (x,t)\in (0,2\pi)\times (0,T],
\end{align}
where $\xi(x,t)$ is a space-time white noise, $f(x,t)=
\alpha(t)e^{-\frac{(x-p)^2}{q^2}}$, $\varepsilon$, $p$, $q$ and
$\sigma$ are model parameters, $\alpha(t)$ is the excitation. Periodic
boundary condition is prescribed and $\varepsilon=0.1$.
%The
%system models heat transfer along a uniform thin rod, and the source
%term $f$ represents an external source centered at $x=p$ with
%time-dependent intensity $\alpha(t)$. 

We conduct the sFML modeling in modal space, i.e., in Fourier
space. Assume the trajectory data of
\eqref{sheat} are observed at grid points $x_j=jh$, $j=0,1,...,N-1$
where $h=2\pi/N$. These observational data are then projected onto a
finite dimensional space $\mathbb{V}_N=\text{span}\{e_k\}_{k=1}^N$, where
$\{e_k\}_{k=1}^\infty$ is an orthogonal basis of
$\mathbb{V}=L^2_{\text{per}}(0,2\pi)$. Thus, the solution
$u(x,t)$ can be approximated by  
\be \label{expansion}
	u_N(x,t) = \sum_{i=0}^N c_k(t)e_k(x),
        \ee
        where we use the Fourier basis for
        $e_k$'s and $c_k$'s are the Fourier coefficients. We solve
\eqref{sheat} with Fourier collocation method with $N=30$.
We then apply the proposed sFML method to learn the dynamical behavior
of the coefficients $\mathbf{c}=[c_0,\dots, c_N]^T$, where the
excitation signal $\alpha(t)$ is represented as 2nd degree Taylor
polynomial with the coefficient $\Gam_0\in
(-1.2,1.2)\times(-3.5,3.5) \times (-5,5)$.
%The training data
%set \eqref{dataset} is generated 
%by sampling $\mathbf{c}_0$ within a proper range and using Taylor
%polynomial of degree $2$ for the control $\alpha(t)$. This introduces
%3 parameters for.
A total of $120,000$
trajectory pairs are used in the training data set \eqref{dataset},
where the time step $\Delta = 0.05$.

For validation, we conduct the sFML simulation for 
$p=1.0$, $q=1.0$,
$\sigma=0.05$, and with $\alpha(t)=\sin t$. The initial condition is set
as $u(x,0)=\exp(-\sin^2 x)-1$. In the top of Figure \ref{fig:SPDE_meanvar}, we
compute the temporal evolution of the mean and standard deviation of
the solution as several locations,
averaged over $10,000$ trajectories. In the bottom of  Figure
\ref{fig:SPDE_meanvar}, we present the solution at several different
time instances $T=0.5, 2.5, 5, 8$. In Figure \ref{fig:SPDE_pdf}, we also show the comparison
of the probability distribution of the solution at a few arbitrarily
chosen spatial-temporal locations, 
$(x,t)=(0.7,2),(1.3,4),(3.8,6),(5.5,8)$. We observe good agreement
between the sFML model prediction and the ground truth.

\begin{figure}[htbp]
  \centering
  \label{fig:SPDE_meanvar}
  \includegraphics[width=.99\textwidth]{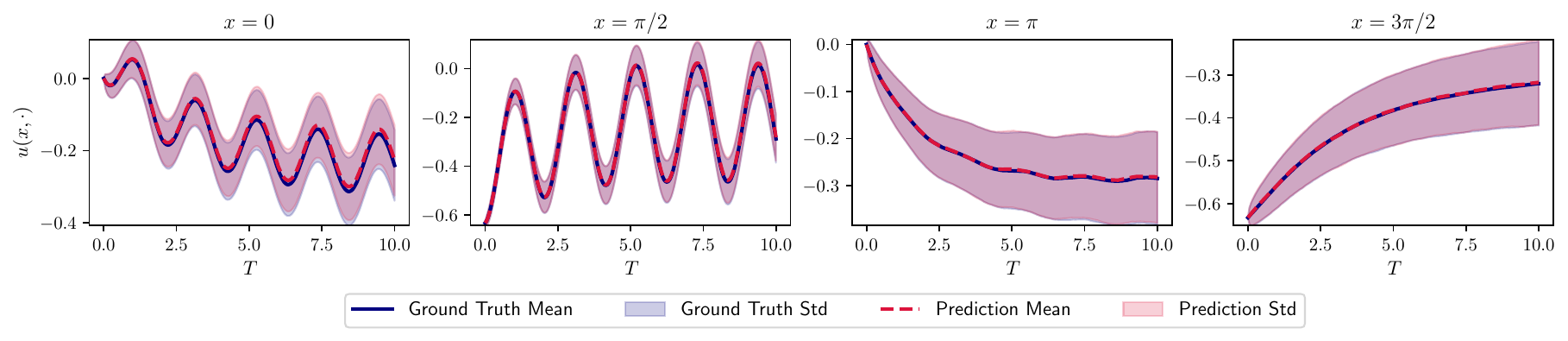}
  \includegraphics[width=.99\textwidth]{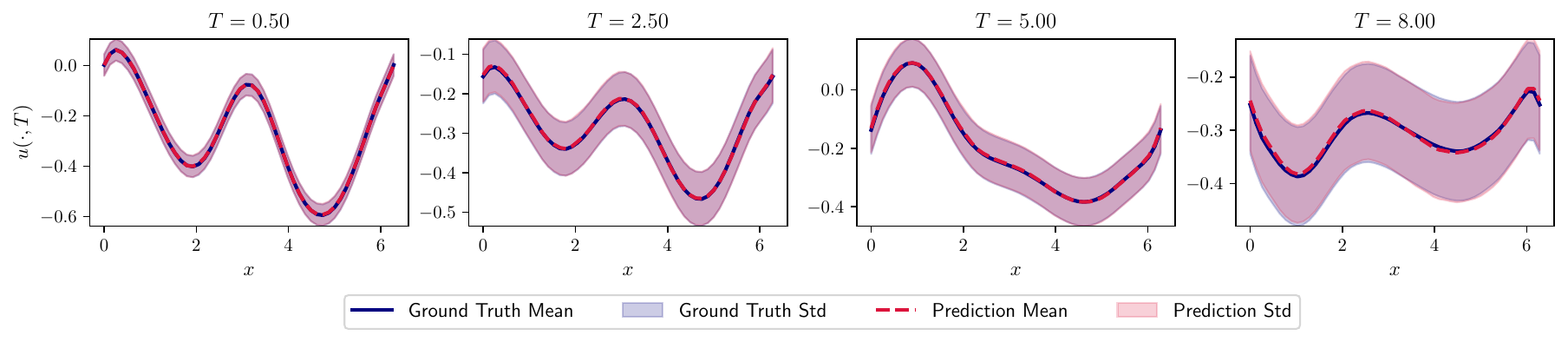}
  \caption{\RV{Mean and standard deviation (STD) of Example \ref{sheat} with initial condition $u(x,0)=\exp(-\sin^2 x)-1$ and $\alpha(t)=\sin t$. Upper: Comparison for $u(x,\cdot)$ with $x=0, \pi/2, \pi, 3\pi/2$. Lower: Comparison for $u(\cdot,T)$ with $T=0.5, 2.5, 5, 8$.}} 
\end{figure}

\begin{figure}[htbp]
  \centering
  \label{fig:SPDE_pdf}
  \includegraphics[width=.99\textwidth]{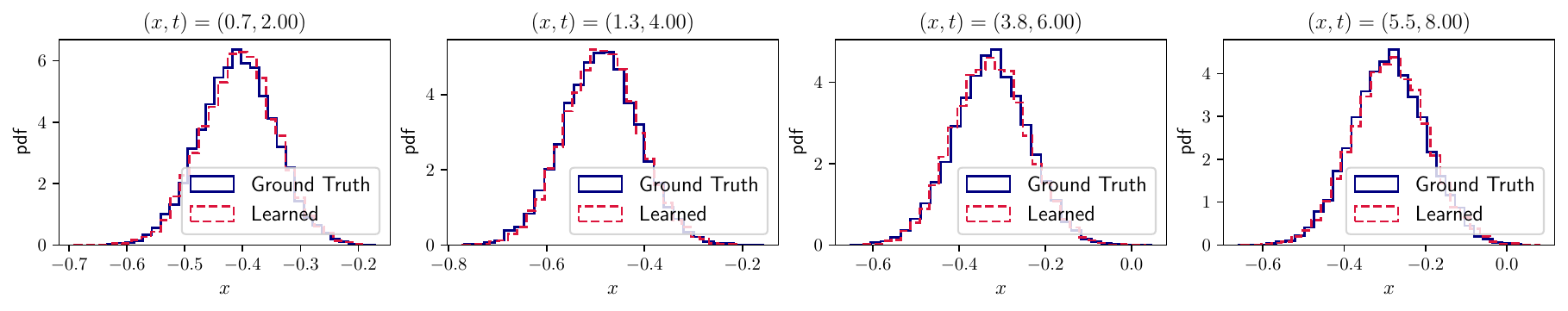}
  \caption{\RV{Comparasion of probability distributions of Example \ref{sheat}  at $(x,t)=(0.7,2),(1.3,4),(3.8,6),(5.5,8)$ with initial condition $u(x,0)=\exp(-\sin^2 x)-1$ and $\alpha(t)=\sin t$.}} 
\end{figure}

}
\section{Conclusion}\label{sec:conclu}
In this paper, we presented a general numerical framework for modeling unknown nonautonomous stochastic systems by using observed trajectory data. To overcome the   difficulties brought by the external time-dependent inputs, we transfer the original system into a local parametric stochastic system. We accomplished this by locally parameterizing the time-dependent external inputs on several discrete time points. The resulting stochastic system is then driven by a stationary parametric stochastic flow map. A normalizing flow model is emplooyed to approximate the parametric stochastic flow map. By using a comprehensive set of numerical examples, we demonstrated that the proposed approach is effective and accurate in modeling a variety of unknown stochastic systems.
\RV{One of the key features of the method is that it is able to produce learned models that can conduct accurate system predictions under external excitations not in the training data and for exceptionally long-term well
  beyond the time domain of the training data.
  Although the method works remarkably well for the variety of examples in the paper, we caution that its rigorous numerical analysis such as error estimate is still lacking, largely due to
  the lack of fudamental results on DNNs. Also, the performance for high dimensional stochastic systems is unclear. These issues are being pursued at this moment and the results will be reported in future studies.}

\bibliographystyle{siamplain}
\bibliography{references}

\begin{thebibliography}{10}

\bibitem{anderson2007modified}
{\sc D.~F. Anderson}, {\em A modified next reaction method for simulating
  chemical systems with time dependent propensities and delays}, The Journal of
  chemical physics, 127 (2007).

\bibitem{pmlr-v1-archambeau07a}
{\sc C.~Archambeau, D.~Cornford, M.~Opper, and J.~Shawe-Taylor}, {\em Gaussian
  process approximations of stochastic differential equations}, in Gaussian
  Processes in Practice, N.~D. Lawrence, A.~Schwaighofer, and
  J.~Quiñonero~Candela, eds., vol.~1 of Proceedings of Machine Learning
  Research, Bletchley Park, UK, 12--13 Jun 2007, PMLR, pp.~1--16,
  \url{https://proceedings.mlr.press/v1/archambeau07a.html}.

\bibitem{benzi1982stochastic}
{\sc R.~Benzi, G.~Parisi, A.~Sutera, and A.~Vulpiani}, {\em Stochastic
  resonance in climatic change}, Tellus, 34 (1982), pp.~10--16.

\bibitem{benzi1983theory}
{\sc R.~Benzi, G.~Parisi, A.~Sutera, and A.~Vulpiani}, {\em A theory of
  stochastic resonance in climatic change}, SIAM J. Appl. Math., 43 (1983),
  pp.~565--478, \url{https://doi.org/10.1137/0143037},
  \url{https://doi.org/10.1137/0143037}.

\bibitem{benzi1981mechanism}
{\sc R.~Benzi, A.~Sutera, and A.~Vulpiani}, {\em The mechanism of stochastic
  resonance}, J. Phys. A, 14 (1981), pp.~L453--L457,
  \url{http://stacks.iop.org/0305-4470/14/L453}.

\bibitem{bowsher2013fidelity}
{\sc C.~G. Bowsher, M.~Voliotis, and P.~S. Swain}, {\em The fidelity of dynamic
  signaling by noisy biomolecular networks}, PLoS computational biology, 9
  (2013), p.~e1002965.

\bibitem{brunton2016discovering}
{\sc S.~L. Brunton, J.~L. Proctor, and J.~N. Kutz}, {\em Discovering governing
  equations from data by sparse identification of nonlinear dynamical systems},
  Proc. Natl. Acad. Sci. USA, 113 (2016), pp.~3932--3937,
  \url{https://doi.org/10.1073/pnas.1517384113}.

\bibitem{brunton2016sparse}
{\sc S.~L. Brunton, J.~L. Proctor, and J.~N. Kutz}, {\em Sparse identification
  of nonlinear dynamics with control (sindyc)}, IFAC-PapersOnLine, 49 (2016),
  pp.~710--715,
  \url{https://doi.org/https://doi.org/10.1016/j.ifacol.2016.10.249},
  \url{https://www.sciencedirect.com/science/article/pii/S2405896316318298}.
\newblock 10th IFAC Symposium on Nonlinear Control Systems NOLCOS 2016.

\bibitem{NEURIPS2018_69386f6b}
{\sc R.~T.~Q. Chen, Y.~Rubanova, J.~Bettencourt, and D.~K. Duvenaud}, {\em
  Neural ordinary differential equations}, in Advances in Neural Information
  Processing Systems, S.~Bengio, H.~Wallach, H.~Larochelle, K.~Grauman,
  N.~Cesa-Bianchi, and R.~Garnett, eds., vol.~31, Curran Associates, Inc.,
  2018,
  \url{https://proceedings.neurips.cc/paper_files/paper/2018/file/69386f6bb1dfed68692a24c8686939b9-Paper.pdf}.

\bibitem{chen2023data}
{\sc X.~Chen, J.~Duan, J.~Hu, and D.~Li}, {\em Data-driven method to learn the
  most probable transition pathway and stochastic differential equation}, Phys.
  D, 443 (2023), pp.~Paper No. 133559, 15,
  \url{https://doi.org/10.1016/j.physd.2022.133559}.

\bibitem{chen2021solving}
{\sc X.~Chen, L.~Yang, J.~Duan, and G.~E. Karniadakis}, {\em Solving inverse
  stochastic problems from discrete particle observations using the
  {F}okker-{P}lanck equation and physics-informed neural networks}, SIAM J.
  Sci. Comput., 43 (2021), pp.~B811--B830,
  \url{https://doi.org/10.1137/20M1360153}.

\bibitem{chen2023learning}
{\sc Y.~Chen and D.~Xiu}, {\em Learning stochastic dynamical system via flow
  map operator}, J. Comput. Phys., 508 (2024), p.~Paper No. 112984,
  \url{https://doi.org/10.1016/j.jcp.2024.112984},
  \url{https://doi.org/10.1016/j.jcp.2024.112984}.

\bibitem{Churchill_2023}
{\sc V.~Churchill and D.~Xiu}, {\em Flow map learning for unknown dynamical
  systems: Overview, implementation, and benchmarks}, Journal of Machine
  Learning for Modeling and Computing, 4 (2023), pp.~173--201.

\bibitem{colbrook2024beyond}
{\sc M.~J. Colbrook, Q.~Li, R.~V. Raut, and A.~Townsend}, {\em Beyond
  expectations: residual dynamic mode decomposition and variance for stochastic
  dynamical systems}, Nonlinear Dynamics, 112 (2024), pp.~2037--2061.

\bibitem{darcy2022one}
{\sc M.~Darcy, B.~Hamzi, G.~Livieri, H.~Owhadi, and P.~Tavallali}, {\em
  One-shot learning of stochastic differential equations with data adapted
  kernels}, Phys. D, 444 (2023), pp.~Paper No. 133583, 18,
  \url{https://doi.org/10.1016/j.physd.2022.133583}.

\bibitem{deng2020modeling}
{\sc R.~Deng, B.~Chang, M.~A. Brubaker, G.~Mori, and A.~Lehrmann}, {\em
  Modeling continuous stochastic processes with dynamic normalizing flows}, in
  Advances in Neural Information Processing Systems, H.~Larochelle, M.~Ranzato,
  R.~Hadsell, M.~Balcan, and H.~Lin, eds., vol.~33, Curran Associates, Inc.,
  2020, pp.~7805--7815,
  \url{https://proceedings.neurips.cc/paper_files/paper/2020/file/58c54802a9fb9526cd0923353a34a7ae-Paper.pdf}.

\bibitem{dietrich2023learning}
{\sc F.~Dietrich, A.~Makeev, G.~Kevrekidis, N.~Evangelou, T.~Bertalan,
  S.~Reich, and I.~G. Kevrekidis}, {\em Learning effective stochastic
  differential equations from microscopic simulations: linking stochastic
  numerics to deep learning}, Chaos, 33 (2023), pp.~Paper No. 023121, 19,
  \url{https://doi.org/10.1063/5.0113632},
  \url{https://doi.org/10.1063/5.0113632}.

\bibitem{dinh2017density}
{\sc L.~Dinh, J.~Sohl-Dickstein, and S.~Bengio}, {\em Density estimation using
  real {NVP}}, in International Conference on Learning Representations, 2017,
  \url{https://openreview.net/forum?id=HkpbnH9lx}.

\bibitem{FuChangXiu_JMLMC20}
{\sc X.~Fu, L.-B. Chang, and D.~Xiu}, {\em Learning reduced systems via deep
  neural networks with memory}, J. Machine Learning Model. Comput., 1 (2020),
  pp.~97--118.

\bibitem{gillespie1977exact}
{\sc D.~T. Gillespie}, {\em Exact stochastic simulation of coupled chemical
  reactions}, The journal of physical chemistry, 81 (1977), pp.~2340--2361.

\bibitem{guo2022normalizing}
{\sc L.~Guo, H.~Wu, and T.~Zhou}, {\em Normalizing field flows: Solving forward
  and inverse stochastic differential equations using physics-informed flow
  models}, Journal of Computational Physics, 461 (2022), p.~111202,
  \url{https://doi.org/https://doi.org/10.1016/j.jcp.2022.111202},
  \url{https://www.sciencedirect.com/science/article/pii/S0021999122002649}.

\bibitem{haarnoja2018latent}
{\sc T.~Haarnoja, K.~Hartikainen, P.~Abbeel, and S.~Levine}, {\em Latent space
  policies for hierarchical reinforcement learning}, in Proceedings of the 35th
  International Conference on Machine Learning, J.~Dy and A.~Krause, eds.,
  vol.~80 of Proceedings of Machine Learning Research, PMLR, 10--15 Jul 2018,
  pp.~1851--1860, \url{https://proceedings.mlr.press/v80/haarnoja18a.html}.

\bibitem{kaiser2018sparse}
{\sc E.~Kaiser, J.~N. Kutz, and S.~L. Brunton}, {\em Sparse identification of
  nonlinear dynamics for model predictive control in the low-data limit},
  Proceedings of the Royal Society A, 474 (2018), p.~20180335.

\bibitem{kang2019identifying}
{\sc S.~H. Kang, W.~Liao, and Y.~Liu}, {\em I{DENT}: identifying differential
  equations with numerical time evolution}, J. Sci. Comput., 87 (2021),
  pp.~Paper No. 1, 27, \url{https://doi.org/10.1007/s10915-020-01404-9}.

\bibitem{Kobyzev2021}
{\sc I.~Kobyzev, S.~Prince, and M.~Brubaker}, {\em Normalizing flows: An
  introduction and review of current methods}, IEEE Trans. Pattern Anal.
  Machine Intel., 43 (2021), pp.~3964--3979.

\bibitem{korda2018linear}
{\sc M.~Korda and I.~Mezi{\'c}}, {\em Linear predictors for nonlinear dynamical
  systems: Koopman operator meets model predictive control}, Automatica, 93
  (2018), pp.~149--160.

\bibitem{laparra2011iterative}
{\sc V.~Laparra, G.~Camps-Valls, and J.~Malo}, {\em Iterative gaussianization:
  From ica to random rotations}, IEEE Transactions on Neural Networks, 22
  (2011), pp.~537--549, \url{https://doi.org/10.1109/TNN.2011.2106511}.

\bibitem{li2021data}
{\sc Y.~Li and J.~Duan}, {\em A data-driven approach for discovering stochastic
  dynamical systems with non-{G}aussian {L}\'{e}vy noise}, Phys. D, 417 (2021),
  pp.~Paper No. 132830, 12, \url{https://doi.org/10.1016/j.physd.2020.132830}.

\bibitem{li2022extracting}
{\sc Y.~Li, Y.~Lu, S.~Xu, and J.~Duan}, {\em Extracting stochastic dynamical
  systems with {$\alpha$}-stable {L}\'{e}vy noise from data}, J. Stat. Mech.
  Theory Exp.,  (2022), pp.~Paper No. 023405, 23,
  \url{https://doi.org/10.1088/1742-5468/ac4e87},
  \url{https://doi.org/10.1088/1742-5468/ac4e87}.

\bibitem{li2020fourier}
{\sc Z.~Li, N.~B. Kovachki, K.~Azizzadenesheli, B.~liu, K.~Bhattacharya,
  A.~Stuart, and A.~Anandkumar}, {\em Fourier neural operator for parametric
  partial differential equations}, in International Conference on Learning
  Representations, 2021, \url{https://openreview.net/forum?id=c8P9NQVtmnO}.

\bibitem{lu2023data}
{\sc H.~Lu and D.~M. Tartakovsky}, {\em Data-driven models of nonautonomous
  systems}, J. Comput. Phys., 507 (2024), p.~Paper No. 112976,
  \url{https://doi.org/10.1016/j.jcp.2024.112976},
  \url{https://doi.org/10.1016/j.jcp.2024.112976}.

\bibitem{lu2022learning}
{\sc Y.~Lu, R.~Maulik, T.~Gao, F.~Dietrich, I.~G. Kevrekidis, and J.~Duan},
  {\em Learning the temporal evolution of multivariate densities via
  normalizing flows}, Chaos, 32 (2022), pp.~Paper No. 033121, 17,
  \url{https://doi.org/10.1063/5.0065093},
  \url{https://doi.org/10.1063/5.0065093}.

\bibitem{mauroy2019koopman}
{\sc A.~Mauroy and J.~Goncalves}, {\em Koopman-based lifting techniques for
  nonlinear systems identification}, IEEE Transactions on Automatic Control, 65
  (2019), pp.~2550--2565.

\bibitem{mezic2004comparison}
{\sc I.~Mezi{\'c} and A.~Banaszuk}, {\em Comparison of systems with complex
  behavior}, Physica D: Nonlinear Phenomena, 197 (2004), pp.~101--133.

\bibitem{muller2019neural}
{\sc T.~M\"{u}ller, B.~Mcwilliams, F.~Rousselle, M.~Gross, and J.~Nov\'{a}k},
  {\em Neural importance sampling}, ACM Trans. Graph., 38 (2019),
  \url{https://doi.org/10.1145/3341156}, \url{https://doi.org/10.1145/3341156}.

\bibitem{oksendal2003stochastic}
{\sc B.~{\O}ksendal}, {\em Stochastic differential equations}, in Stochastic
  differential equations, Springer, 2003, pp.~65--84.

\bibitem{opper2019variational}
{\sc M.~Opper}, {\em Variational inference for stochastic differential
  equations}, Ann. Phys., 531 (2019), pp.~1800233, 9,
  \url{https://doi.org/10.1002/andp.201800233}.

\bibitem{otto2021koopman}
{\sc S.~E. Otto and C.~W. Rowley}, {\em Koopman operators for estimation and
  control of dynamical systems}, Annual Review of Control, Robotics, and
  Autonomous Systems, 4 (2021), pp.~59--87.

\bibitem{owhadi2021computational}
{\sc H.~Owhadi}, {\em Computational graph completion}, Research in the
  Mathematical Sciences, 9 (2022), p.~27.

\bibitem{papamakarios2021normalizing}
{\sc G.~Papamakarios, E.~Nalisnick, D.~J. Rezende, S.~Mohamed, and
  B.~Lakshminarayanan}, {\em Normalizing flows for probabilistic modeling and
  inference}, J. Mach. Learn. Res., 22 (2021), pp.~Paper No. 57, 64.

\bibitem{Papamakarios21}
{\sc G.~Papamakarios, E.~Nalisnick, D.~J. Rezende, S.~Mohamed, and
  B.~Lakshminarayanan}, {\em Normalizing flows for probabilistic modeling and
  inference}, J. Machine Learning Res., 22 (2021), pp.~1--64.

\bibitem{NIPS2017_6c1da886}
{\sc G.~Papamakarios, T.~Pavlakou, and I.~Murray}, {\em Masked autoregressive
  flow for density estimation}, in Advances in Neural Information Processing
  Systems, I.~Guyon, U.~V. Luxburg, S.~Bengio, H.~Wallach, R.~Fergus,
  S.~Vishwanathan, and R.~Garnett, eds., vol.~30, Curran Associates, Inc.,
  2017,
  \url{https://proceedings.neurips.cc/paper_files/paper/2017/file/6c1da886822c67822bcf3679d04369fa-Paper.pdf}.

\bibitem{proctor2016dynamic}
{\sc J.~L. Proctor, S.~L. Brunton, and J.~N. Kutz}, {\em Dynamic mode
  decomposition with control}, SIAM J. Appl. Dyn. Syst., 15 (2016),
  pp.~142--161, \url{https://doi.org/10.1137/15M1013857},
  \url{https://doi.org/10.1137/15M1013857}.

\bibitem{proctor2018generalizing}
{\sc J.~L. Proctor, S.~L. Brunton, and J.~N. Kutz}, {\em Generalizing {K}oopman
  theory to allow for inputs and control}, SIAM J. Appl. Dyn. Syst., 17 (2018),
  pp.~909--930, \url{https://doi.org/10.1137/16M1062296},
  \url{https://doi.org/10.1137/16M1062296}.

\bibitem{qin2021data}
{\sc T.~Qin, Z.~Chen, J.~D. Jakeman, and D.~Xiu}, {\em Data-driven learning of
  nonautonomous systems}, SIAM J. Sci. Comput., 43 (2021), pp.~A1607--A1624,
  \url{https://doi.org/10.1137/20M1342859}.

\bibitem{QinChenJakemanXiu_IJUQ}
{\sc T.~Qin, Z.~Chen, J.~D. Jakeman, and D.~Xiu}, {\em Deep learning of
  parameterized equations with applications to uncertainty quantification},
  Int. J. Uncertain. Quantif., 11 (2021), pp.~63--82,
  \url{https://doi.org/10.1615/Int.J.UncertaintyQuantification.2020034123},
  \url{https://doi.org/10.1615/Int.J.UncertaintyQuantification.2020034123}.

\bibitem{qin2019data}
{\sc T.~Qin, K.~Wu, and D.~Xiu}, {\em Data driven governing equations
  approximation using deep neural networks}, J. Comput. Phys., 395 (2019),
  pp.~620--635, \url{https://doi.org/10.1016/j.jcp.2019.06.042}.

\bibitem{raissi2019physics}
{\sc M.~Raissi, P.~Perdikaris, and G.~Karniadakis}, {\em Physics-informed
  neural networks: A deep learning framework for solving forward and inverse
  problems involving nonlinear partial differential equations}, Journal of
  Computational Physics, 378 (2019), pp.~686--707,
  \url{https://doi.org/10.1016/j.jcp.2018.10.045}.

\bibitem{raissi2018multistep}
{\sc M.~Raissi, P.~Perdikaris, and G.~E. Karniadakis}, {\em Multistep neural
  networks for data-driven discovery of nonlinear dynamical systems}, arXiv
  preprint arXiv:1801.01236,  (2018).

\bibitem{schaeffer2017sparse}
{\sc H.~Schaeffer and S.~G. McCalla}, {\em Sparse model selection via integral
  terms}, Phys. Rev. E, 96 (2017), pp.~023302, 7,
  \url{https://doi.org/10.1103/physreve.96.023302}.

\bibitem{schaeffer2018extracting}
{\sc H.~Schaeffer, G.~Tran, and R.~Ward}, {\em Extracting sparse
  high-dimensional dynamics from limited data}, SIAM J. Appl. Math., 78 (2018),
  pp.~3279--3295, \url{https://doi.org/10.1137/18M116798X}.

\bibitem{song2017nice}
{\sc J.~Song, S.~Zhao, and S.~Ermon}, {\em A-nice-mc: Adversarial training for
  mcmc}, in Advances in Neural Information Processing Systems, I.~Guyon, U.~V.
  Luxburg, S.~Bengio, H.~Wallach, R.~Fergus, S.~Vishwanathan, and R.~Garnett,
  eds., vol.~30, Curran Associates, Inc., 2017,
  \url{https://proceedings.neurips.cc/paper_files/paper/2017/file/2417dc8af8570f274e6775d4d60496da-Paper.pdf}.

\bibitem{voliotis2016stochastic}
{\sc M.~Voliotis, P.~Thomas, R.~Grima, and C.~G. Bowsher}, {\em Stochastic
  simulation of biomolecular networks in dynamic environments}, PLoS
  computational biology, 12 (2016), p.~e1004923.

\bibitem{wang2022data}
{\sc Y.~Wang, H.~Fang, J.~Jin, G.~Ma, X.~He, X.~Dai, Z.~Yue, C.~Cheng, H.-T.
  Zhang, D.~Pu, D.~Wu, Y.~Yuan, J.~Gonçalves, J.~Kurths, and H.~Ding}, {\em
  Data-driven discovery of stochastic differential equations}, Engineering, 17
  (2022), pp.~244--252,
  \url{https://doi.org/https://doi.org/10.1016/j.eng.2022.02.007}.

\bibitem{weinan2011principles}
{\sc E.~Weinan}, {\em Principles of multiscale modeling}, Cambridge University
  Press, 2011.

\bibitem{xu2023learning}
{\sc Z.~Xu, Y.~Chen, Q.~Chen, and D.~Xiu}, {\em Modeling unknown stochastic
  dynamical system via autoencoder}, arXiv preprint arXiv:2312.10001,  (2023).

\bibitem{yang2022generative}
{\sc L.~Yang, C.~Daskalakis, and G.~E. Karniadakis}, {\em Generative ensemble
  regression: Learning particle dynamics from observations of ensembles with
  physics-informed deep generative models}, SIAM Journal on Scientific
  Computing, 44 (2022), pp.~B80--B99, \url{https://doi.org/10.1137/21M1413018}.

\bibitem{yildiz2018learning}
{\sc C.~Yildiz, M.~Heinonen, J.~Intosalmi, H.~Mannerstrom, and H.~Lahdesmaki},
  {\em Learning stochastic differential equations with gaussian processes
  without gradient matching}, in 2018 IEEE 28th International Workshop on
  Machine Learning for Signal Processing (MLSP), IEEE, 2018, pp.~1--6.

\bibitem{YuEtal_2022}
{\sc L.~Yu, R.~Maulik, T.~Gao, F.~Dietrich, I.~G. Kevrekidis, and J.~Duan},
  {\em Learning the temporal evolution of multivariate densities via
  normalizing flows}, Chaos, 32 (2022).

\bibitem{zhang2022multiauto}
{\sc J.~Zhang, S.~Zhang, and G.~Lin}, {\em Multiauto-deeponet: A
  multi-resolution autoencoder deeponet for nonlinear dimension reduction,
  uncertainty quantification and operator learning of forward and inverse
  stochastic problems}, arXiv preprint arXiv:2204.03193,  (2022).

\bibitem{zhu2024dyngma}
{\sc A.~Zhu and Q.~Li}, {\em Dyngma: a robust approach for learning stochastic
  differential equations from data}, arXiv preprint arXiv:2402.14475,  (2024).

\end{thebibliography}
\end{document}